\documentclass[journal]{IEEEtran}
\usepackage{graphicx}
\newcommand{\fref}[1]{Fig. \ref{#1}}
\newcommand{\sref}[1]{Section \ref{#1}}
\newcommand{\tref}[1]{TABLE \ref{#1}}

\usepackage{array}
\usepackage{amsmath}
\usepackage{bm}

\usepackage{algorithm}
\usepackage{algorithmic}

\usepackage{array}
\usepackage{multirow}
\usepackage{booktabs}
\usepackage[table,xcdraw]{xcolor}
\usepackage{graphicx}
\ifCLASSOPTIONcompsoc
\usepackage[caption=false, font=normalsize, labelfont=sf, textfont=sf]{subfig}
\else
\usepackage[caption=false, font=footnotesize]{subfig}
\fi
\usepackage{url}
\usepackage{color,xcolor}
\usepackage{hyperref}

\begin{document}

\title{Hybridization of evolutionary algorithm and deep reinforcement learning for multi-objective orienteering optimization}

%

\author{Wei~Liu,
        Rui~Wang,
        Tao~Zhang,
        Kaiwen~Li,
        Wenhua~Li,
        and Hisao Ishibuchi, \emph{Fellow}, \emph{IEEE}

\thanks{This work was supported by the National Science Fund for Outstanding Young Scholars (62122093), the Scientific key Research Project of National University of Defense Technology (ZZKY-ZX-11-04) and the National Natural Science Foundation of China (72071205). \emph{(Corresponding authors: Rui~Wang; Tao~Zhang.)}} 

\thanks{Wei~Liu, Rui~Wang, Tao~Zhang, Kaiwen~Li and Wenhua~Li are with the College of System Engineering, National University of Defense Technology, Changsha 410073, PR China, and with the Hunan Key Laboratory of Multi-Energy System Intelligent Interconnection Technology, HKL-MSI2T, Changsha 410073, PR China (e-mail: weiliu16@nudt.edu.cn, ruiwangnudt@gmail.com, zhangtao@nudt.edu.cn, kaiwenli\_nudt@foxmail.com, liwenhua1030@aliyun.com).}

\thanks{Hisao Ishibuchi is with Guangdong Provincial Key Laboratory of Braininspired Intelligent Computation, and Department of Computer Science and Engineering, Southern University of Science and Technology, Shenzhen 518055, China (e-mail: hisao@sustech.edu.cn).}
}

%
%


\maketitle

\begin{abstract}
Multi-objective orienteering problems (MO-OPs) are classical multi-objective routing problems and have received a lot of attention in the past decades. This study seeks to solve MO-OPs through a problem-decomposition framework, that is, a MO-OP is decomposed into a multi-objective knapsack problem (MOKP) and a travelling salesman problem (TSP). The MOKP and TSP are then solved by a multi-objective evolutionary algorithm (MOEA) and a deep reinforcement learning (DRL) method, respectively. While the MOEA module is for selecting cities, the DRL module is for planning a Hamiltonian path for these cities. An iterative use of these two modules drives the population towards the Pareto front of MO-OPs. The effectiveness of the proposed method is compared against NSGA-II and NSGA-III on various types of MO-OP instances. Experimental results show that our method exhibits the best performance on almost all the test instances, and has shown strong generalization ability.
\end{abstract}


\IEEEpeerreviewmaketitle

\section{Introduction}
Orienteering problem (OP), one of the most classical combinatorial optimization problems (COPs), arises regularly in real world. Typically, an OP is described as follows. While there are a number of cities with different locations and profits, a decision-maker needs to select some cities and determine the shortest tour for the selected cities such that the total profit collected from the visited cities is maximized. Since the first introduction of OP by Golden et al. \cite{golden1987orienteering} in 1987, a number of algorithms have been proposed to deal with OPs \cite{laporte1990selective,ramesh1992optimal,leifer1994strong}. In addition, several variants of OP \cite{vansteenwegen2011orienteering,gunawan2016orienteering} have been introduced such as time dependent orienteering problems (TDOPs), team orienteering problems (TOPs), orienteering problems with time windows (OPTWs) and multi-objective orienteering problems (MO-OPs). These problems have received increasing attentions in recent years. 

This study focuses on MO-OPs. A typical application scenario of MO-OPs is the tourism route planning, in which each scenic spot has multiple profits (for example, profits in culture and entertainment). The objectives of MO-OPs are maximizing the total profits in multiple dimensions and minimizing the total tour length. Since MO-OPs are NP-hard, exact methods become unaffordable when the problem scale is large. Alternatively, metaheuristics, in particular, multi-objective evolutionary algorithms (MOEAs) \cite{deb2002fast,zhang2007moea} become the mainstream for dealing with MO-OPs. To name a few, Schilde et al. \cite{schilde2009metaheuristics} combined Path Relinking procedures into two MOEAs, i.e., a Pareto ant colony optimization algorithm and an extended variable neighborhood search method. Experimental results on both benchmark instances and real-world instances demonstrated their good performance. In addition, Marti et al. combined the GRASP and Path Relinking \cite{marti2015multiobjective}; Malt et al. proposed a large neighborhood search (LNS) method \cite{matl2017bi}. Martin et al. \cite{martin2018multi} developed a multi-objective artificial bee colony (MOABC) algorithm; Bossek et al. \cite{bossek2018local} integrated local search techniques into NSGA-II. Mei et al. \cite{mei2016efficient} proposed a multi-objective memetic algorithm (MOMA) and a multi-objective ant colony system (MACS) algorithm for a time-dependent MO-OP.

Although MOEAs have been widely applied to solve MO-OPs as well as other multi-objective optimization problems (MOPs), they have shown certain limitations \cite{zhang2016decision,ming2019evolutionary}. First, MOEAs are a kind of iteration-based methods which have to evolve for a large number of generations to find the global optimal solutions, especially in large-scale instances. This results in a long running time. Second, recombination operators in MOEAs have to be designed carefully, which requires rich domain knowledge in both MOEAs and MOPs. Furthermore, such algorithms are usually optimized for specific tasks, which means that a slightly change of the problem may require to re-design the algorithm.

In recent years, deep reinforcement learning (DRL) has shown promising performance in combinatorial optimization problems \cite{kai2021research}. Bello et al. \cite{bello2016neural} used a reinforcement learning (RL) algorithm to train a pointer network (PN) \cite{vinyals2015pointer} for solving travelling salesman problems (TSPs). Nazari et al. \cite{nazari2018reinforcement} simplified the encoder module of the PN and introduced dynamic information as a part of inputs, making the model suitable for vehicle routing problems (VRPs). Kool et al. \cite{kool2018attention} proposed an attention model to solve the routing problems, including OPs. Li et al. \cite{li2020deep} proposed a DRL-MOA framework to solve MOPs by DRL algorithms. Experimental results in multi-objective TSP instances showed its competitiveness in terms of both model performance and running time. 

The main advantage of solving combinatorial optimization problems by DRL methods is that the DRL model can generate solutions immediately once trained, since it is an end-to-end method. Also, the generalization ability of DRL methods is attractive. The trained model is applicable to a set of similar instances with different scales \cite{li2020deep}. Despite these advantages, the training of a DRL model costs much in terms of both time and computing resource. Also, it faces difficulty to solve complex problems involving a large number of decision variables such as large-scale MO-OPs. Moreover, whereas the application of DRL to combinatorial optimization problems has been examined in many studies, its performance is still poor on some problems.

Therefore, it is difficult for existing methods to effectively solve large-scale MO-OPs. For a complex problem, decomposing it into several subproblems and solving these subproblems collaboratively is usually a good idea. In this way, not only the complexity of the original problem gets reduced, but also the algorithms designed for each subproblem are more targeted. However, how to reasonably decompose the original problem and design effective algorithms for the subproblems remain challenging.

Since an OP essentially can be decomposed into two subproblems, i.e., city selection and Hamiltonian path planning \cite{vansteenwegen2011orienteering}, it is possible to solve OPs more effectively through problem decomposition. In this study, we propose to decompose a MO-OP into two basic combinatorial optimization problems, i.e., MOKP and TSP, and propose a hybrid optimization framework, namely, MOEA-DRL, to tackle MO-OPs. In MOEA-DRL, we treat the city selection as a MOKP, and solve it using a MOEA. We treat the Hamiltonian path planning as a TSP, and solve it using a pre-trained DRL model. Once the DRL model is trained, it can immediately generate a solution (city visit sequence) for the cities selected by the MOKP solver, and feed the solution back to the MOKP solver, helping it adjust the selection of cities. Updating solutions iteratively eventually solves the given MO-OP. The global search ability of MOEAs and the powerful generalization ability of DRL models make the MOEA-DRL framework effective, being applicable to large-scale MO-OPs. Specifically, in the MOEA-DRL framework NSGA-II and NSGA-III are used as the MOKP solver (i.e., two variants of MOEA-DRL are implemented), and an improved PN is used as the TSP solver. The performance of MOEA-DRL is evaluated by comparing it with NSGA-II and NSGA-III. These two MOEAs are implemented for MO-OPs and applied under various specifications of the termination conditions (i.e., the number of generations).

It is worth mentioning that in the MOEA-DRL framework DRL is much more suitable to serve as the TSP solver than other algorithms, for example, evolutionary algorithms (EAs), commercial optimization solvers (for example, Gurobi \cite{optimization2018gurobi}) and local search methods (e.g., 2-opt method). This is because that the TSP solver has to be called many times during evolution while these algorithms are all iteration-based. Different from them, as an end-to-end method, a DRL model can output optimal solutions immediately. In addition, it is clear that the optimization quality of TSP greatly impacts the final performance of MOEA-DRL on MO-OPs. While 2-opt is usually effective, it was found inferior to well-trained DRL models on TSPs, especially for large-scale instances \cite{kool2018attention,li2020deep}. These are the reasons why we use a DRL model as the TSP solver in the MOEA-DRL framework. We have also tried to improve it and conduct adequate pre-training of the DRL model. The main contributions are summarized as follows:

\begin{itemize}
\item A hybrid optimization framework for dealing with MO-OPs, namely, MOEA-DRL, is proposed. Due to the modularity and simplicity of the MOEA-DRL framework, it is possible to use any suitable MOEA and DRL algorithms. The MOEA-DRL framework inherits the global search ability of MOEAs and the strong generalization ability of DRL, and performs well on a set of MO-OP instances.
\item The conventional pointer network is improved. Except for static information in the model inputs, distance information is embedded as a dynamic embedding, helping with city decoding. The introduction of a dynamic embedding mechanism effectively speeds up the convergence during model training.
\item The performance of the proposed MOEA-DRL framework is evaluated through computational experiments on various MO-OP instances (i.e., bi-objective instances and three-objective instances) with different problem size. The MOEA-DRL framework shows clear advantages with respect to optimization quality, especially for large-scale instances.
\item High generalization ability of the MOEA-DRL framework is demonstrated. When the DRL module is trained using 100-city instances, the MOEA-DRL performs well on MO-OP instances ranging from 20 cities to 1000 cities.
\end{itemize}

The rest of this study is organized as follows. \sref{model} first formulates the mathematical model of MO-OPs, and then illustrates the proposed MOEA-DRL framework, including MOEAs and a dynamic pointer network. After that, the setting of computational experiments is described in \sref{setup}, and the experimental results are analyzed in \sref{results}. \sref{conclusion} concludes this study and identifies some future research directions.

\section{Proposed Model}
\label{model}

\subsection{Problem Formulation}
\label{problem}

A MO-OP is formulated as the following $(K+1)$-objective problem where a subset $S$ of the given city set $V$ is selected and a tour for the cities in $S$ is determined:

\begin{equation}\label{eq:objects}
\text { Maximize } f(S)=\left(f_{1}(S), \ldots, f_{K}(S),-f_{K+1}(S)\right),
\end{equation}
subject to
\begin{equation}\label{eq:constraint1}
v_{1} \in S \subseteq V, 
\end{equation}
\begin{equation}\label{eq:constraint2}
\emph{TourLength}(S) \leq T_{\max },
\end{equation}
where 
\begin{equation}\label{eq:object1}
f_{k}(S)=\sum_{v_{i} \in S} s_{i}^{k}, k=1,2, \ldots, K,
\end{equation}

\begin{equation}\label{eq:object2}
f_{K+1}(S)=\emph{TourLength}(S).
\end{equation}

In this formulation, $s_i^k$ is the $k$th profit from city $v_i$, $\emph{TourLength}(S)$ is the tour length for the cities in $S$, and $T_{max}$ is the upper bound for the tour length. It should be noted that the calculation of $\emph{TourLength}(S)$ needs a TSP optimizer whereas the total profits in \eqref{eq:object1} are easily calculated. Thus, the MO-OP in \eqref{eq:objects}-\eqref{eq:object2} can be viewed as a bi-level multi-objective optimization problem. In the outer level, multi-objective city selection is performed. Then, the tour length is minimized for the selected cities in the inner level. In this study, we propose to handle the optimal city selection by an MOEA, and handle the tour length minimization by a DRL algorithm.

\subsection{General Framework of MOEA-DRL}
\label{Framework}
Without the preference of decision makers, multi-objective evolutionary algorithms are the most common approaches for MOPs, and they can be classified into three categories, i.e., Pareto dominance-based, decomposition-based, and indicator-based. For Pareto dominance-based MOEAs, e.g, NSGA-II, solutions are evaluated based on the Pareto dominance relation as well as an additional diversity metric. Decomposition-based MOEAs decompose a MOP into a set of subproblems by weighted scalarizing methods, and solve these subproblems in a collaborative manner. Representatives include Cellular-based MOGA \cite{murata2001specification}, MOEA/D, MOEA/DD \cite{li2014evolutionary} and NSGA-III \cite{deb2013evolutionary}. Indicator-based MOEAs search for the best solution set by optimizing a performance indicator such as hypervolume (HV) \cite{fonseca2006improved} and inverted generational distance (IGD) \cite{sun2018igd}. The proposed framework MOEA-DRL can use any of MOEAs, specifically depending on the choice of the MOKP solver.

In the proposed hybrid optimization framework, MOEA-DRL, a MO-OP is decomposed into a MOKP and a TSP. The MOKP is solved by MOEAs, resulting in a set of selected cities. Each individual in the MOEA presents a plan of city selection. Based on the selected cities, a DRL model is used as the TSP solver to output a Hamiltonian path. The obtained tour length for each individual (i.e., each selection plan of cities) is fed back to the MOEA and used in the constraint condition and the $(K+1)$th objective in \eqref{eq:object2}. Non-dominated solutions in the final population are the optimization results obtained by the MOEA-DRL where each solution has the corresponding Hamiltonian path. The general framework of MOEA-DRL is presented in Algorithm \ref{alg:MOEA-DRL}. As described in lines 2 and 3, the model training and population initialization are required before the process of evolution. 

\begin{algorithm}[htb]
\caption{General Framework of MOEA-DRL}
\label{alg:MOEA-DRL}
\begin{algorithmic}[1]
\REQUIRE {instance set $\mathcal{M}$, maximum number of generations ${MaxGen}$, population size $N$}
\STATE Initialize ${gen} \leftarrow 1$
\STATE Train DRL model for TSP: $DRLModel \leftarrow REINFORCE(\mathcal{M})$
\STATE $Pop \leftarrow Initialization(N)$
\WHILE{${gen} \leq {MaxGen}$}
\STATE $Route \leftarrow DRLModel(Pop)$
\STATE $Pop \leftarrow MOEA(Pop,Route,N)$
\STATE $gen \leftarrow gen+1$
\ENDWHILE
\STATE $Route \leftarrow DRLModel(Pop)$
\STATE $Arc \leftarrow UpdateArc(Pop,Route)$
\end{algorithmic}
\end{algorithm}

Compared with pure MOEAs, the problem decomposition strategy can reduce the complexity of the problem effectively. Since the MOEA module in MOEA-DRL only focuses on city selection, it can be coded through binary variables, which is much simpler than the traditional permutation coding method for MO-OPs. The reduction of problem complexity makes EAs efficient and applicable to large-scale instances. Compared with pure DRL models, the training of the DRL model in the MOEA-DRL framework is much easier. This is because in this framework the DRL model is designed for TSP with less complexity than OP. It only needs the geographical information about the city locations. Thus, once trained, it is applicable to MO-OPs with any city profit distribution, tour length limitation and multiple objectives. Since the training of the DRL model is time-consuming, these advantages are vital for its application to practical problems. Moreover, any MOEA can be employed as the MOKP solver and any end-to-end method for TSP \cite{kool2018attention,gama2021reinforcement} can be integrated as the TSP solver in the MOEA-DRL framework. This framework is applicable to various MO-OP variants by simply adding related constraints to MOKP and TSP solvers.

In the MOEA-DRL framework, we test NSGA-II and NSGA-III as MOKP solvers, and introduce a dynamic pointer network (DYPN) as the TSP solver. These will be described in detail in the following subsections.

\subsection{MOEAs for MOKP}
\label{MOEA}
Since we need to select cities first, this subproblem is modeled as a MOKP. Among various approaches proposed for MOKPs in literature, MOEAs showed promising performance \cite{zitzler1999multiobjective}. NSGA-II and NSGA-III are well-known and frequently-used MOEAs and are embedded into our framework as MOKP solvers. Following the general coding method of MOKPs, the binary coding method is used in the MOEA-DRL framework.

Considering that a good initial population can accelerate the convergence, and improve the solution quality of evolutionary algorithms, this study proposes a greedy heuristic for population initialization on MO-OP instances. For each individual, cities are selected step by step based on an indicator, namely, the profit density. Assuming that the last visited city is $i$, the profit density vector $(I^{i}_{1},I^{i}_{2},\ldots,I^{i}_{N})$ is defined as the ratio of the profit vector $(s_{1},s_{2},\ldots,s_{N})$ and the Euclidean distance vector $(e_{i1},e_{i2},\ldots,e_{iN})$ ($e_{i j}$ denotes the Euclidean distance between the city $i$ and $j$). It is formulated as follows:

\begin{equation}
I^{i}_{j}=s_{j}/e_{i j}, \quad i=1, \ldots, N; j=1, \ldots, N (i\neq j),
\label{eq:indicator}
\end{equation}
where $s_{j}$ is calculated as the sum:
\begin{equation}
s_{j}=\sum_{k=1}^{K} s_{j}^{k}.
\label{eq:profit}
\end{equation}

The indicator $I^{i}_{j}$ in \eqref{eq:indicator} shows how worthy city $j$ is when the last selected city is $i$. The higher the value of $I^{i}_{j}$ for city $j$ is, the more likely $j$ is to be selected in that step. Since each city can be visited only once, all the visited cities should be excluded from the next city selection procedure. The next city is then sampled based on a probability vector $P^{i}=(P^{i}_{j}\mid j \in U_i)$, which is the normalization form of $I^{i}_{U_{i}}=(I^{i}_{j}\mid j \in U_i)$ as follows: 

\begin{equation}
P^{i}=\operatorname{softmax}(I^{i}_{U_{i}}) , \quad i=1, \ldots, N,
\label{eq:probs}
\end{equation}
where $U_i$ is the set of unvisited cities when the last visited city is $i$. 

\subsection{Dynamic Pointer Network for TSP}
\label{DYPN}
Since the TSP solver is called many times, its computational efficiency is necessary. DRL is suitable for this purpose due to its end-to-end model property. High performance of DRL on TSPs has been repeatedly demonstrated in literature. Nazari et al. \cite{nazari2018reinforcement} improved the PN \cite{vinyals2015pointer} by introducing an attention mechanism and verified its effect on TSP and VRP. Furthermore, in this work, based on the model of Nazari, we propose a dynamic pointer network (DYPN) for TSP, by introducing dynamic information into the model.

Taking a TSP instance with $N$ cities as an example, the input and output of DYPN are the city locations $X=\{x_i \mid i=1,\ldots,N\}$ and the city permutation $\boldsymbol{\pi}=\left(\pi_{1}, \dots, \pi_{N}\right)$, respectively. Following the probability chain rule, the policy $p(\boldsymbol{\pi} \mid r)$ for giving a solution $\boldsymbol{\pi}$ on case $r$ can be defined as \eqref{eq:prob chain rule} where $\theta$ represents the parameters to be learned. 
\begin{equation}
p_{\theta}(\boldsymbol{\pi} \mid r)=\prod_{t=1}^{N} p_{\theta}\left(\pi_{t} \mid r, {\pi}_{0},\ldots,{\pi}_{t-1}\right).
\label{eq:prob chain rule}
\end{equation}

Here, $\pi_{0}$ presents the initial state, that is, no city has been visited. At each decoding step $t=1,\ldots,N$, city $\pi_{t}$ is selected from available cities based on the probability $p_{\theta}\left(\pi_{t}\mid r,{\pi}_{0},\ldots,{\pi}_{t-1}\right)$. It does not matter whether the depot was selected first or not, since the tour of all the selected cities is always a circle in a TSP. The decoding step of the depot makes no difference on the total tour length.

\begin{figure}[htbp]
	\centering
	\includegraphics[width=1\linewidth]{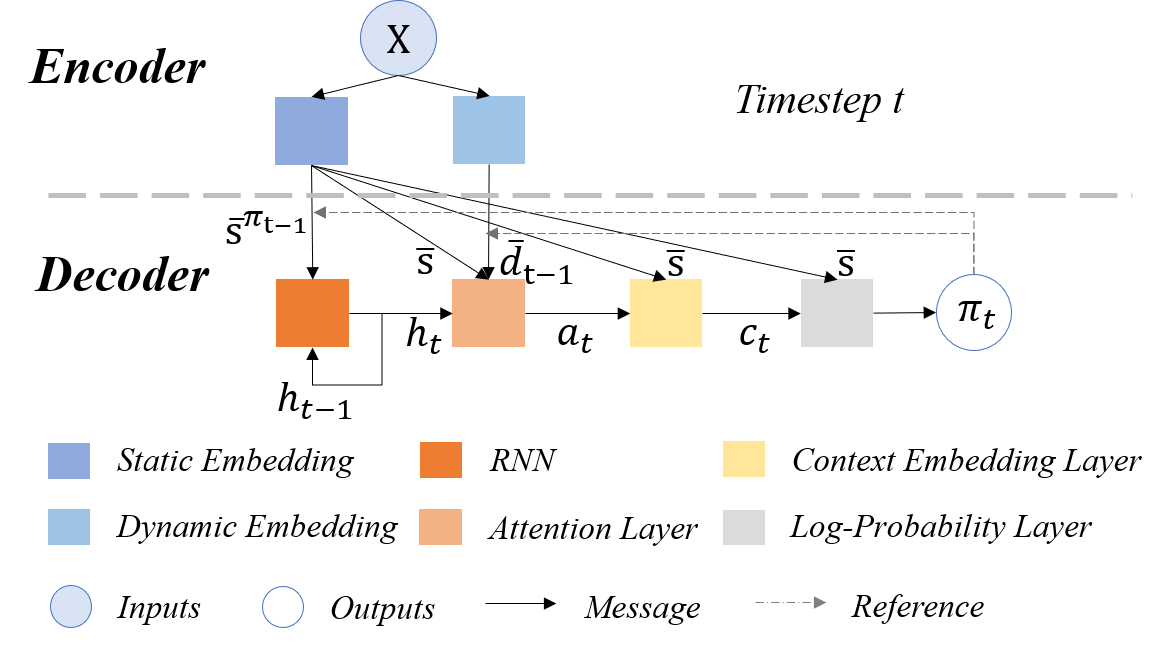}
	\caption{Illustration of the model structure. The model takes the location information as inputs, and it outputs the city sequence step by step. Static embedding and dynamic embedding are included in the encoder. An RNN, an attention layer, a context embedding layer and a log-probability layer constitute the decoder.}
	\label{fig:model}
\end{figure}

The DYPN follows the encoder-decoder architecture as many other DRL models. The encoder is used to map the inputs into a high-dimensional knowledge vector, and the decoder is used to decode it to a desired sequence. The structure of DYPN is shown in \fref{fig:model}.

\subsubsection{Encoder}
\label{encoder}
The encoder is structured to condense the input information into a knowledge vector. Following the design in \cite{nazari2018reinforcement}, the 1-dimensional (1-D) convolution layer is used as the encoder in this work. As the sequential information of the TSP inputs is meaningless, complex encoders like RNN \cite{bello2016neural} and multi-head attention (MHA) \cite{kool2018attention,gama2021reinforcement,li2021deep} mechanisms are unnecessary here. In each 1-D convolution layer, the number of in-channels is the dimension of inputs, and the number of out-channels is set as $d_h$ ($d_h$=128 in this study). There are two 1-D convolution layers being introduced in this study: one for static embedding, and the other for dynamic embedding. The parameters of each 1-D convolution layer are shared among all the static inputs or the dynamic inputs. The inputs are defined as $X=\{x^{i} \mid i=1,\dots,N\}$, in which each input $x^{i}$ is formed as a set of tuples $\{x_{t}^{i}=\left(s^{i},d_{t}^{i} \right) \mid t=0,\dots,T\}$. $s^{i}$ and $d_{t}^{i}$ are the static feature and the dynamic feature at step $t$, respectively. 

In the static encoder, since the city location is a 2-dimensional vector, the number of the in-channels of the 1-D convolution layer is set to 2. Thus, the static encoder maps a $N*2$ vector to a $N*d_h$ vector (suppose there are $N$ cities). This is the static embedding, and it is formed as $\bar{s}=\{\bar{s}^{i} \mid i=1,\ldots,N\}$.

Considering that the distance information between cities is important in planning a path, we introduce an Euclidean distance vector $E_{t}=(e_{v_t 1},e_{v_t 2},\dots,e_{v_t N})$ ($e_{v_t j}$ denotes the Euclidean distance between the city $j$ and the city $v_t$, which is the city selected at decoding step $t$) as the dynamic feature. Before the dynamic encoding, data normalization is applied to speed up the model learning, as formulated in \eqref{eq:normalization} where $E_{t}^{max}$ and $E_{t}^{min}$ are the maximum and minimum values among the elements of the vector $E_{t}$.

\begin{equation}
d_{t}=\left\{\frac{E_{t}^{max}-e_{v_t i}}{E_{t}^{max}-E_{t}^{min}}\mid i=1, \ldots, N\right\}.
\label{eq:normalization}
\end{equation}

Similar to the process of the static embedding, the dynamic feature $d_t$ is mapped by a 1-D convolution layer from the size of $N*1$ to $N*d_h$. The embedded dynamic vector in step $t$ is formulated as $\bar{d}_{t}$. As described in \fref{fig:model}, the dynamic embedding $\bar{d}_{t}$ is only used in the attention layer.

\subsubsection{Decoder}
\label{decoder}

The decoder is used to decode the embeddings into a city sequence. In this work, the decoder is composed of a RNN, an attention layer, a context embedding layer and a log-probability layer, as shown in \fref{fig:model}. The RNN summarizes the information of the visited cities; the attention layer measures how important each city is in a decoding step; the context embedding layer generates a weighted static embedding; and the log-probability layer is used to calculate a probability vector for city selection.

As illustrated in \fref{fig:model}, the decoding process works as follows. In timestep $t\in\{1,\dots,N\}$, the RNN (a gated recurrent unit (GRU) \cite{cho2014learning} in this study) takes the static embedding $\bar{s}^{\pi_{t-1}}$ of the last decoded city and its last memory state $h_{t-1}$ (if $t>1$) as inputs, and it outputs a new memory state $h_{t}$, as formulated in \eqref{eq:GRU} below. Since no city has been selected before step $t=1$, $\bar{s}^{\pi_{0}}$ in \eqref{eq:GRU} is initialized by the embedding of a zero vector with the size of $2*1$ . 

\begin{equation}
h_{t}=f_{G R U}\left(\bar{s}^{\pi_{t-1}}, h_{t-1}\right).
\label{eq:GRU}
\end{equation}

The attention layer then takes the memory state $h_{t}$, static embedding $\bar{s}$ of all cities and the dynamic embedding $\bar{d}_{t}$ at step $t$ as inputs. An attention vector $a_{t}$ is calculated as shown in \eqref{eq:attention1} and \eqref{eq:attention2} below. $v_{a}$ and $W_{a}$ in \eqref{eq:attention1} are learnable parameters, and ``$;$" represents the concatenation of two vectors. 

\begin{equation}
u_{t}=v_{a}^{T} \tanh \left(W_{a}\left[\bar{s};\bar{d}_{t}; h_{t}\right]\right),
\label{eq:attention1}
\end{equation}

\begin{equation}
a_{t}=\operatorname{softmax}\left(u_{t}\right).
\label{eq:attention2}
\end{equation}

After that, the context embedding layer is used to produce a weighted static embedding vector $c_{t}$ as follows:

\begin{equation}
c_{t}= a_{t} \bar{s}.
\label{eq:context}
\end{equation}

Finally, the probability vector mentioned in \eqref{eq:prob chain rule} is calculated as follows:

\begin{equation}
\tilde{u}_{t}=v_{c}^{T} \tanh \left(W_{c}\left[\bar{s} ; c_{t}\right]\right),
\label{eq:prob1}
\end{equation}

\begin{equation}
p_{\theta}\left(\pi_{t} \mid r, {\pi}_{0},\ldots,{\pi}_{t-1}\right)=\operatorname{softmax}\left(\tilde{u}_{t}\right),
\label{eq:prob2}
\end{equation}
where $v_{c}$ and $W_{c}$ are learnable parameters. 

Based on the probability vector, the city to be visited is selected by an inference method \cite{gama2021reinforcement}. In this work, the sampling search strategy is used in the training stage to increase the exploration ability, and the greedy search strategy is used in the validation stage. While the sampling search strategy samples one city based on the probability vector $p_{\theta}\left(\pi_{t} \right)$ in each step $t$, the greedy search strategy directly selects the city with the highest probability. The selected cities are masked to avoid repetitive visit. The order of the selected cities in the $N$ decoding steps creates a solution $(\pi_{1}, \pi_{2}, \ldots,\pi_{N} )$ of the MO-OP.

\subsubsection{Training Method}
\label{train}
In this study, the REINFORCE algorithm \cite{williams1992simple} is used as a RL method to train the DYPN model. There are two networks to be trained: an actor network and a critic network. While the actor provides policies for determining the next action, the critic evaluates the reward of the given policy. For a TSP instance $r$, the actor and the critic are parameterized by $\theta$ and $\phi$, respectively. The reward of the policy $\boldsymbol{\pi}$ given by the actor is formulated as $L(\boldsymbol{\pi})$, and the evaluated reward given by the critic is formulated as $V(r;\phi)$. The loss function is defined as \eqref{eq:loss}, and its gradient is formed as \eqref{eq:gradient loss}.

\begin{equation}
\mathcal{L}({\theta} \mid r)={E}_{p_{{\theta}}(\boldsymbol{\pi} \mid r)}[L(\boldsymbol{\pi})-V(r;\phi)],
\label{eq:loss}
\end{equation}

\begin{equation}
\nabla \mathcal{L}({\theta} \mid r)={E}_{p_{{\theta}}(\boldsymbol{\pi} \mid r)}\left[(L(\boldsymbol{\pi})-V(r;\phi)) \nabla \log p_{{\theta}}(\boldsymbol{\pi} \mid r)\right].
\label{eq:gradient loss}
\end{equation}

The actor network used here is DYPN, and the critic is constructed as a multi-layer dense network. The parameter settings for the actor and the critic are listed in \tref{table:network parameters}. The training process is described in Algorithm \ref{alg:reinforce}. We first initialize the actor and critic networks with random parameters $\theta$ and $\phi$ in [-1, 1], respectively. In each training epoch, $M$ instances are drawn from a problem set $\mathcal{M}$. For each instance $r_{m}$, the actor provides a solution $\boldsymbol{\pi}$. Then the reward $L^{m}(\boldsymbol{\pi})$ and the evaluated reward $V(r_{m};\phi)$ are calculated. After the rewards and the approximated rewards are calculated for all $M$ instances, the parameters of the actor and the critic are updated. Such a parameter update process is iterated for $N_{epoch}$ times.

\begin{table*}[ht]
\centering
\caption{Parameter settings of the DRL model}
\begin{tabular}{ccc}
\hline 
\multicolumn{3}{c}{\cellcolor[HTML]{FFFFFF}\textbf{Actor Network}} \\
\hline 
\textbf{Module} & \textbf{Type} & \textbf{Parameters}\\ 
\hline 
Encoder & 1D-Conv & $D_{input}=D_{problem},\ D_{output}=128,\ kernel\ size=1,\ stride=1$ \\ 
Decoder & GRU & $D_{input} = 128,\ D_{output} = 128,\ hidden\ size=128,\ number\ of\ layer=1$ \\ 
 & Other layers & $No \ hyper-parameters$ \\ 
\hline 
\multicolumn{3}{c}{\cellcolor[HTML]{FFFFFF}\textbf{Critic Network}} \\
\hline 
\textbf{Module} & \textbf{Type} & \textbf{Parameters}\\ 
\hline 
First layer &  1D-Conv & $D_{input}=2*hidden\ size,\ D_{output}=20,\ kernel\ size=1,\ stride=1$ \\
Second layer &  1D-Conv & $D_{input}=20,\ D_{output}=20,\ kernel\ size=1,\ stride=1$ \\
Third layer &  1D-Conv & $D_{input}=20,\ D_{output}=1,\ kernel\ size=1,\ stride=1$ \\
\hline 
\end{tabular} 
\label{table:network parameters}
\end{table*}

\begin{algorithm}[htb]
\caption{REINFORCE training algorithm}
\label{alg:reinforce}
\begin{algorithmic}[1]
\REQUIRE {problem set $\mathcal{M}$, number of instances $M$ for each training epoch, number of training epochs $N_{epoch}$}
\STATE Initialize the actor network with random weights $\theta$ and the critic network with random weights $\phi$
\FOR{$epoch = 1:N_{epoch}$}
\STATE reset gradients: $d \theta \leftarrow 0, d \phi \leftarrow 0$
\STATE sample $M$ instances from the problem set $\mathcal{M}$
\FOR{instance $r_{m} = r_{1},\ldots,r_{M}$}
\FOR{step $t = 1:N_{step}$}
\STATE ${\pi}_{t} \leftarrow p_{{\theta}}(r_{m},{\pi}_{0},\ldots,{\pi}_{t-1}).$
\ENDFOR
\STATE compute reward $L^{m}(\boldsymbol{\pi})$
\STATE compute estimated reward $V(r_{m};\phi)$
\ENDFOR
\STATE $d \theta \leftarrow \frac{1}{M} \sum_{m=1}^{M}\left(L^{m}(\boldsymbol{\pi})-V(r_{m};\phi)\right) \nabla_{{\theta}} \log p_{{\theta}}(\boldsymbol{\pi} \mid r_{m})$

\STATE $d \phi \leftarrow \frac{1}{M} \sum_{m=1}^{M}\nabla_{\boldsymbol{\phi}} \left(L^{m}(\boldsymbol{\pi})-V(r_{m};\phi)\right)^{2}$

\STATE Update ${\theta}$ using $d \theta$ and $\boldsymbol{\phi}$ using $d \phi$
\ENDFOR
\end{algorithmic}
\end{algorithm}

\section{Experimental Setup}
\label{setup}

All experiments in this study are conducted on a single RTX 3060 GPU. The code is written in Python 3.8 and will be open access once the paper was accepted for the convenience of experimental reproduction and further research. All competitor algorithms are implemented with the help of the open library Geatpy\footnote{https://github.com/geatpy-dev/geatpy}, which provides various EAs. This section illustrates the instance setup, hyper-parameter settings and the evaluation procedure.

\subsection{Instance Setup}
\label{instance}

\textbf{Training set:}
As a machine learning method, the DRL model DYPN needs to be trained in advance. It is worth noting that its training requires only the location information of cities since it is designed for the TSP module. This is the case in any type of MO-OP instances. Different from the MOEA-DRL framework, the training of pure DRL models designed for OPs or MO-OPs \cite{kool2018attention,li2020deep} requires not only the location information but also the city profits and the total tour length constraint. This means that any change of the city profits or the total tour length constraint in new instances may lead to performance deterioration or failure of the model. In this work, 1,280,000 instances of 100-city TSP (i.e., 1,280,000 locations of 100 cities) are randomly generated as the training set. The locations are all generated in a unit square [0, 1]*[0, 1] with the random seed of 1234.

\textbf{Test set:}
Bi-objective OP instances (i.e., profits type instances and mixed type instances) as well as three-objective instances are considered in this study. The profits type instance has two profits in each city \cite{schilde2009metaheuristics}, and the two objectives are defined by the sum profits of the selected cities. The mixed type instance has two objectives--- to maximize the total single-criterion profit and to minimize the total tour length \cite{bederina2017hybrid}. The three-objective instance has the two total profit objectives and the total tour length objective.

To create test instances, the location of each city, one or two profits at each city, and a total tour length constraint are needed. The generation of the city profits also obeys the uniform distribution over [0, 1]. With the random seed of 12345, 20-, 50-, 100-, 200-, 500- and 1000-city test instances are generated. The total tour length constraint is set as 2, 3, 4, 6, 10, 15 in the 20-, 50-, 100-, 500- and 1000-city test instances, respectively. HV is used as a performance indicator in this work. The reference point for calculating HV values is set as (0, -2), (0, -3), (0, -4), (0, -6), (0, -10) and (0, -15) for mixed type test instances with 20, 50, 100, 200, 500 and 1000 cities, respectively. Similarly, the reference point for the three-objective instances is (0, 0, -2), (0, 0, -3), (0, 0, -4), (0, 0, -6), (0, 0, -10) and (0, 0, -15) for each problem size. The reference point corresponding to the profits type test instances is set as (0, 0) for all instances.

\subsection{Hyper-parameters}
\label{parameters}
The hyper-parameters used in the actor and critic networks are listed in \tref{table:network parameters}. 1D-Conv means the 1-D convolution layer. $D_{input}$, $D_{output}$ and $D_{problem}$ respectively mean the dimension of inputs, outputs and the problem data. Specifically, $D_{problem}=2$ for the static one and $D_{problem}=1$ for the dynamic one in the encoders. The parameters in the training and testing phases are listed in \tref{table:parameters}. Both the actor and critic networks are trained by the Adam optimizer \cite{kingma2014adam} with the learning rate of 0.0001 and the dropout rate of 0.1 for 10 epochs, each of which contains 1,280,000 training instances. The training batch size is set as 64 constrained by the memory limitation. As for the parameters in the testing stage, the population size and the maximum number of generations are set as 100 and 20, respectively. Specific parameters in genetic operators follow the default settings in Geatpy.

\begin{table}[htbp]
\setlength\tabcolsep{3pt}
\caption{Parameter settings of training and testing}
\begin{tabular}{cc|cc}
\hline 
\multicolumn{2}{c|}{\cellcolor[HTML]{FFFFFF}\textbf{Training Stage}} & \multicolumn{2}{c}{\cellcolor[HTML]{FFFFFF}\textbf{Testing stage}} \\
\hline 
\textbf{Hyper-parameters} & \textbf{Value} & \textbf{Hyper-parameters} & \textbf{Value}\\ 
\hline 
Number of epochs & 10 & Population size & 100 \\ 
Number of instances & 1,280,000 & Max number of generations & 20 \\ 
Batch size & 64 & Probability of crossover & Default \\ 
Optimizer & Adam& Probability of mutation & Default \\ 
Dropout rate& 0.1 \\
Learning rate & 1e-4 &  \\ 

\hline 
\end{tabular} 
\label{table:parameters}
\end{table}

\subsection{Evaluation Procedure}
\label{Evaluation}
In the proposed MOEA-DRL framework, NSGA-II and NSGA-III are respectively examined as MOKP solvers, and DYPN is used as TSP solver. To evaluate the effectiveness of MOEA-DRL, NSGA-II and NSGA-III are also directly used to solve MO-OPs. The maximum number of generations is set as 500, 2000, 10000 and 40000 in each of these two MOEAs (i.e., four different termination conditions are examined for comparison). The open library Geatpy of python is used for the implementation of NSGA-II and NSGA-III. In NSGA-II and NSGA-III, genetic operators are used in the recommended settings in Geatpy. Two different coding methods of compared MOEAs for a MO-OP are examined--- a single-chromosome based permutation coding method \cite{kim2020multi} and a double-chromosome coding method \cite{bossek2018local}. 

\begin{figure}[htbp]
	\centering
	\includegraphics[width=1\linewidth]{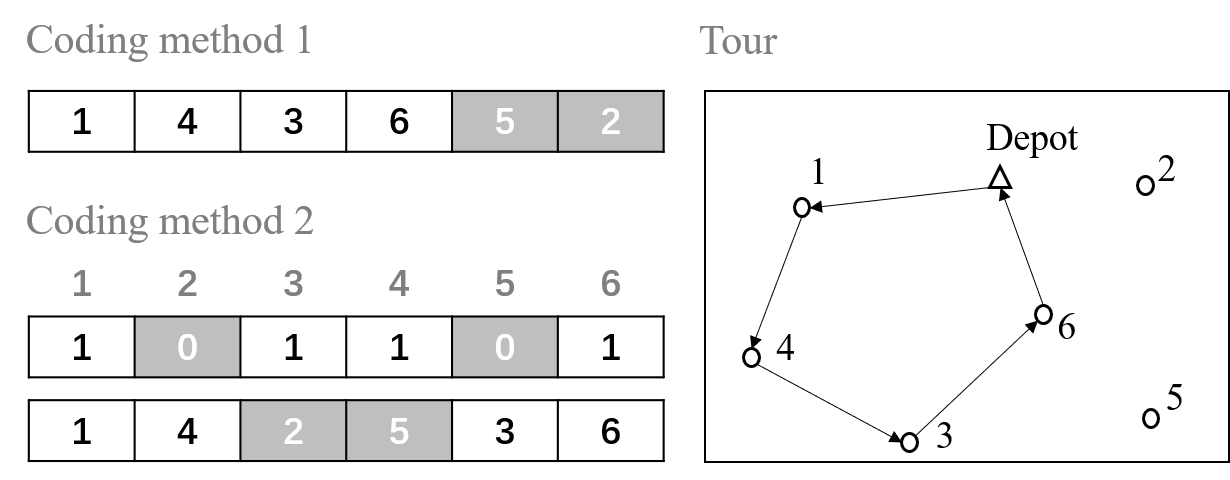}
	\caption{Two coding methods for a MO-OP. Selected cities are shown in white background.}
	\label{fig:Chrom}
\end{figure}

As the depot is always fixed, it is not included in chromosomes. The single-chromosome permutation coding method first permutes the other cities. Following this permutation, these cities are visited as long as the total tour length constraint is not violated. The double-chromosome coding method contains two chromosomes for city selection and city permutation, respectively. The first chromosome uses binary coding and the second chromosome uses permutation coding. A coding example of these two methods is shown in \fref{fig:Chrom}. The two coding examples in \fref{fig:Chrom} generate the same tour $[Depot,\ 1,\ 4,\ 3,\ 6,\ Depot]$.

\section{Experimental Results and Discussion}
\label{results}
\subsection{Training of DRL Module}
\label{training}
\begin{figure}[ht]
	\centering
	\includegraphics[width=0.8\linewidth]{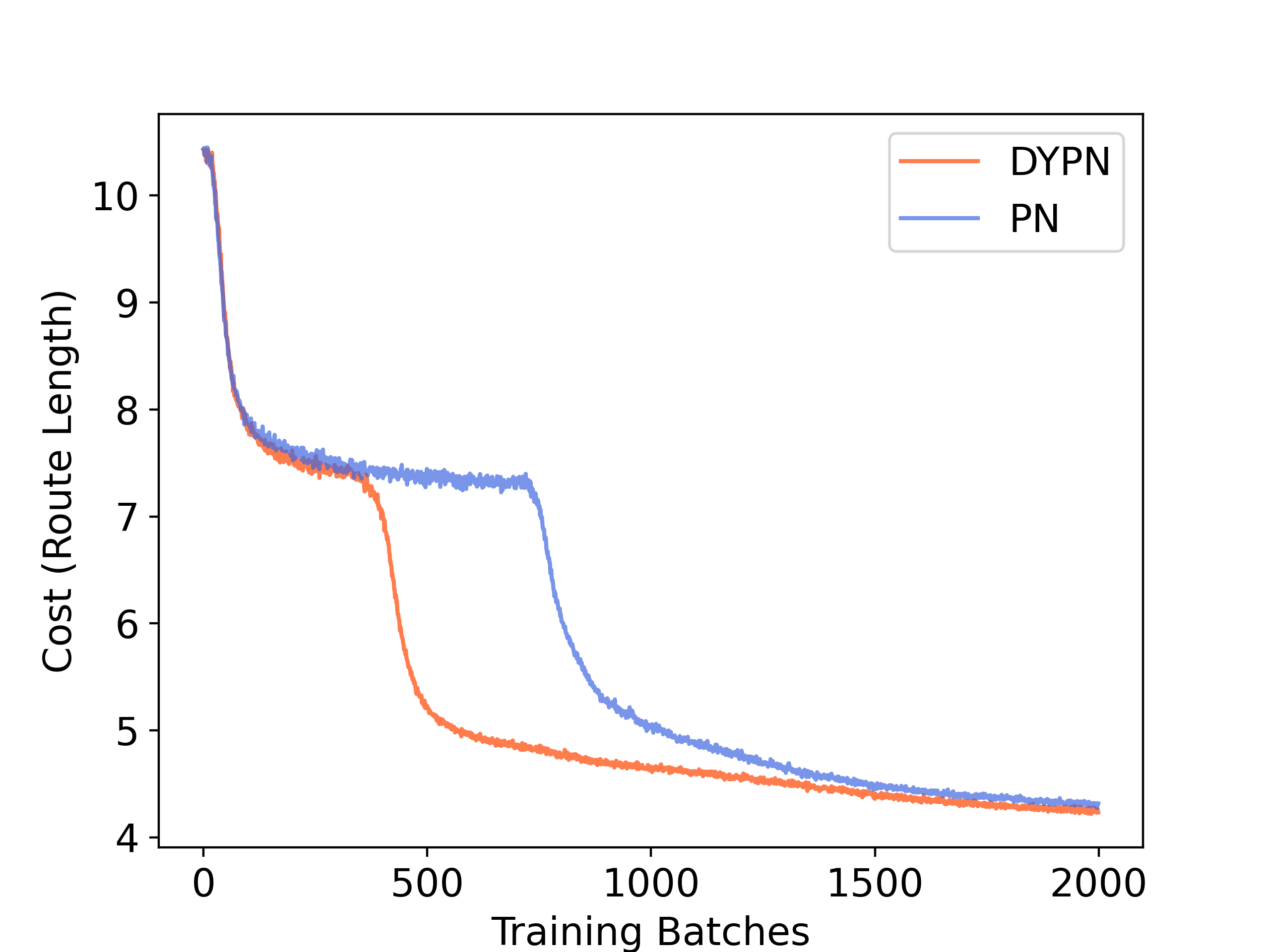}
	\caption{Cost changes during 2000 training batches of DYPN and PN.}
	\label{fig:training cost}
\end{figure}

As a deep network method, DYPN requires a large number of instances for model training, and the training process is time-consuming. Dynamic embedding is introduced in this study to help with the learning and speed up the convergence. In this subsection, we compare DYPN with PN \cite{nazari2018reinforcement} on the training stage. Taking the model training on 20-city TSP instances as an example, the number of training batches is set as 2000 and the batch size is 1024. Each of the 100 training batches take 40.8 and 37.8 seconds on DYPN and PN, respectively. Costs of DYPN and PN during the training are presented in \fref{fig:training cost}. The cost is defined by the tour length to be minimized. In \fref{fig:training cost}, while PN needs about 1000 training batches to converge to the cost of 5, DYPN needs about 500 training batches. In addition to the improvement on learning speed, the optimization ability of DYPN is also stronger than PN even when both of them have been trained for 2000 batches. DYPN trained on 100-city TSP instances is used as a TSP solver in the MOEA-DRL framework for all testing instances from 20 cities to 1000 cities. When training it on 1,280,000 instances for 10 epochs (the training batch size is 64), the total training time of DYPN is about 18 hours.

\subsection{Effectiveness of Two Coding Methods}
\label{coding methods}
\begin{figure}[htbp]
	\centering
	\includegraphics[width=1\linewidth]{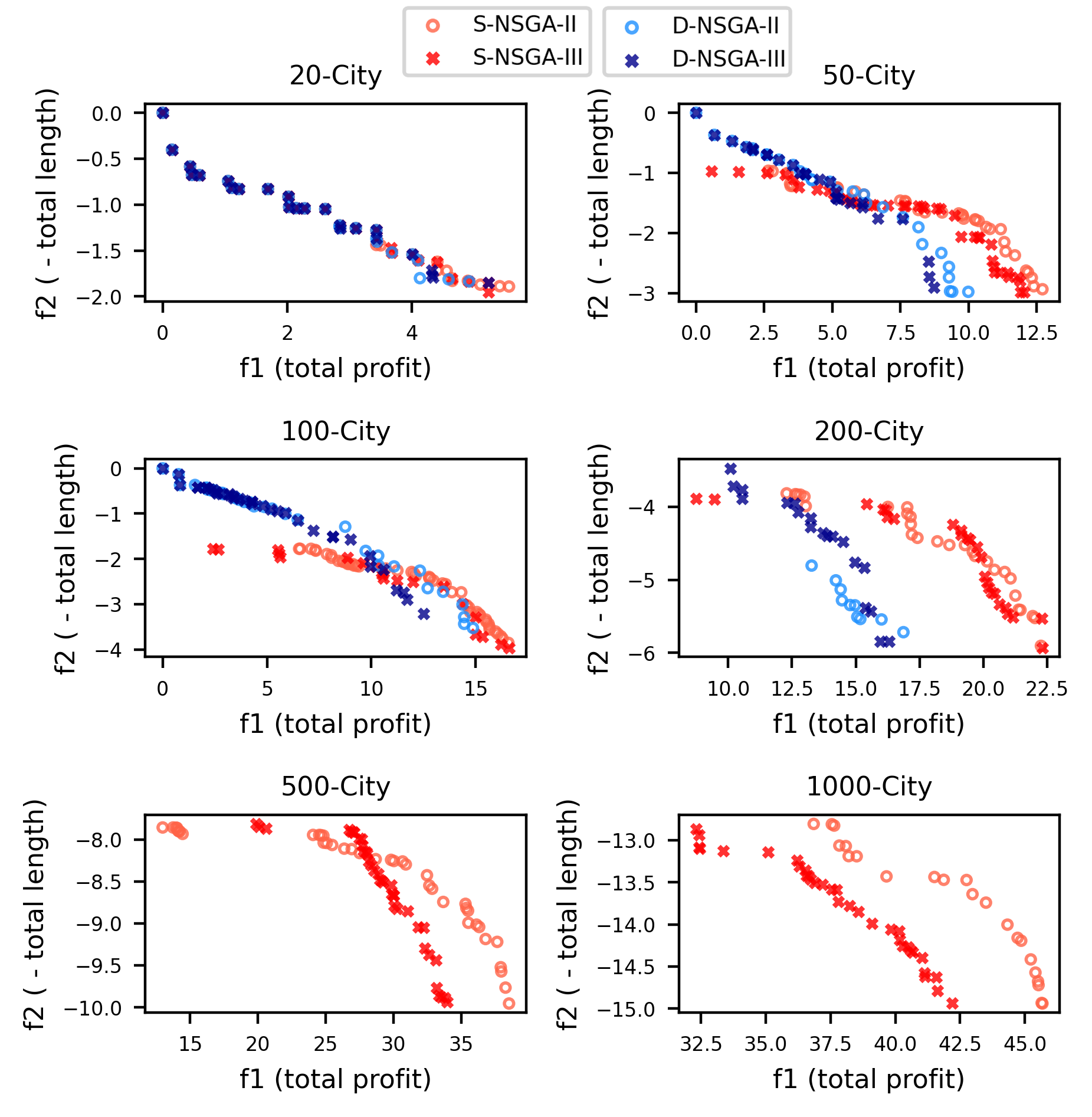}
	\caption{Solution sets obtained by NSGA-II and NSGA-III based on the single-chromosome and double-chromosome coding methods. The test instances are \textbf{Mixed type} bi-objective OP instances with 20, 50, 100, 200, 500 and 1000 cities. Algorithms with the double-chromosome coding method lose efficacy in 500- and 1000-city cases.}
	\label{fig:coding method1}
\end{figure}

\begin{figure}[htbp]
	\centering
	\includegraphics[width=1\linewidth]{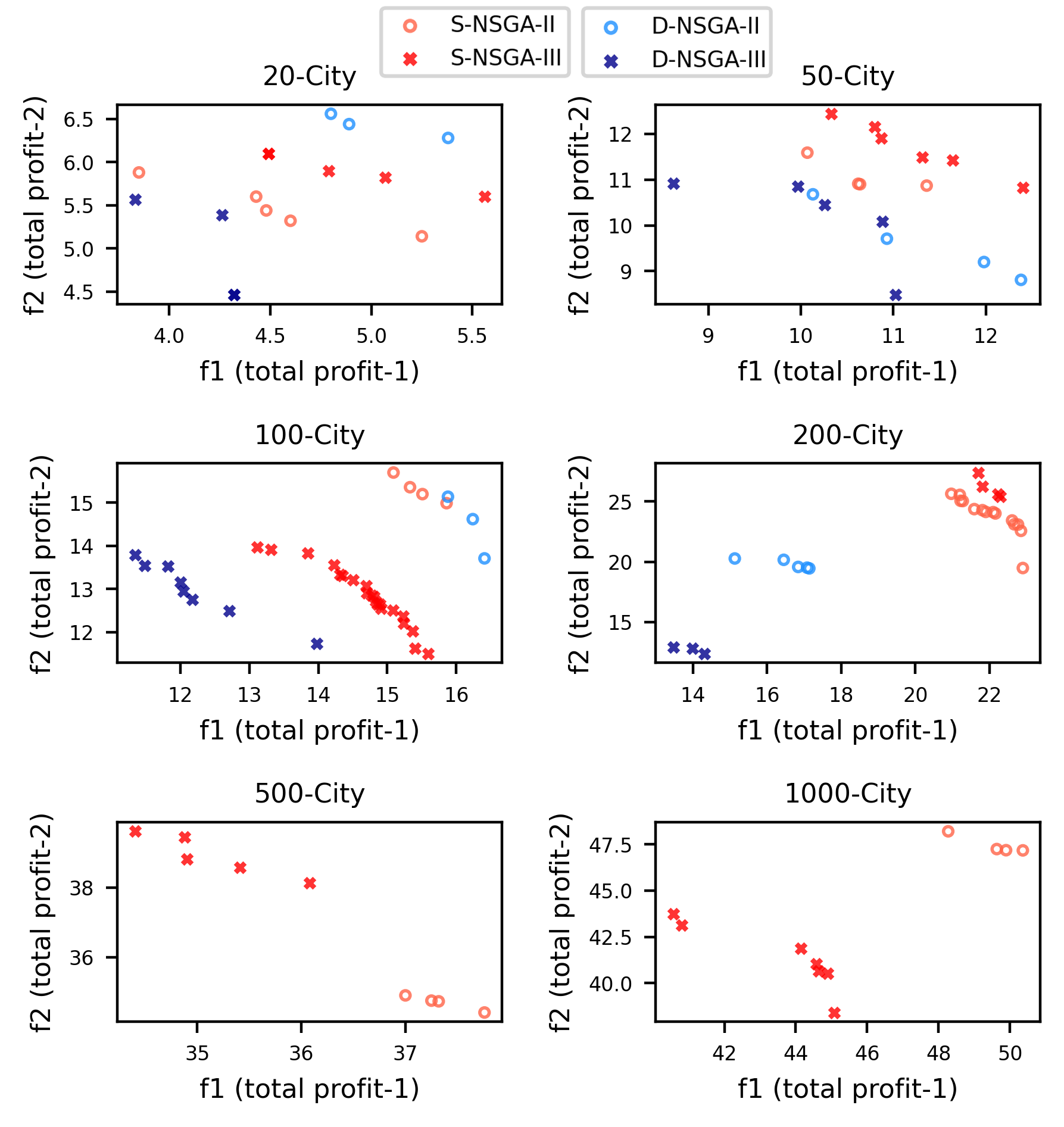}
	\caption{Solution sets obtained by NSGA-II and NSGA-III based on the single-chromosome and double-chromosome coding methods. The test instances are \textbf{Profits type} bi-objective OP instances with 20, 50, 100, 200, 500 and 1000 cities. Algorithms with the double-chromosome coding method lose efficacy in 500- and 1000-city cases.}
	\label{fig:coding method2}
\end{figure}

The single-chromosome permutation coding method and the double-chromosome coding method are two frequently-used coding methods of MO-OPs. They are compared on bi-objectives OPs using NSGA-II and NSGA-III (thus called ``S-NSGA-II", ``S-NSGA-II", ``D-NSGA-III", and ``D-NSGA-III", respectively) in this subsection. Experiments are conducted on 20-, 50-, 100-, 200-, 500- and 1000-city instances. The test instances include both profits type and mixed type instances, which are generated in the same manner as described previously in \sref{instance}. The population size and the maximum number of generations are set as 100 and 500, respectively, in both NSGA-II and NSGA-III. 31 runs of each algorithm are conducted on each test instance. \fref{fig:coding method1} and \fref{fig:coding method2} show the obtained solution set of a single run with the median performance from those 31 runs. 

\begin{figure}
\centering
\subfloat[Mixed type instances]{\includegraphics[width=1.7in]{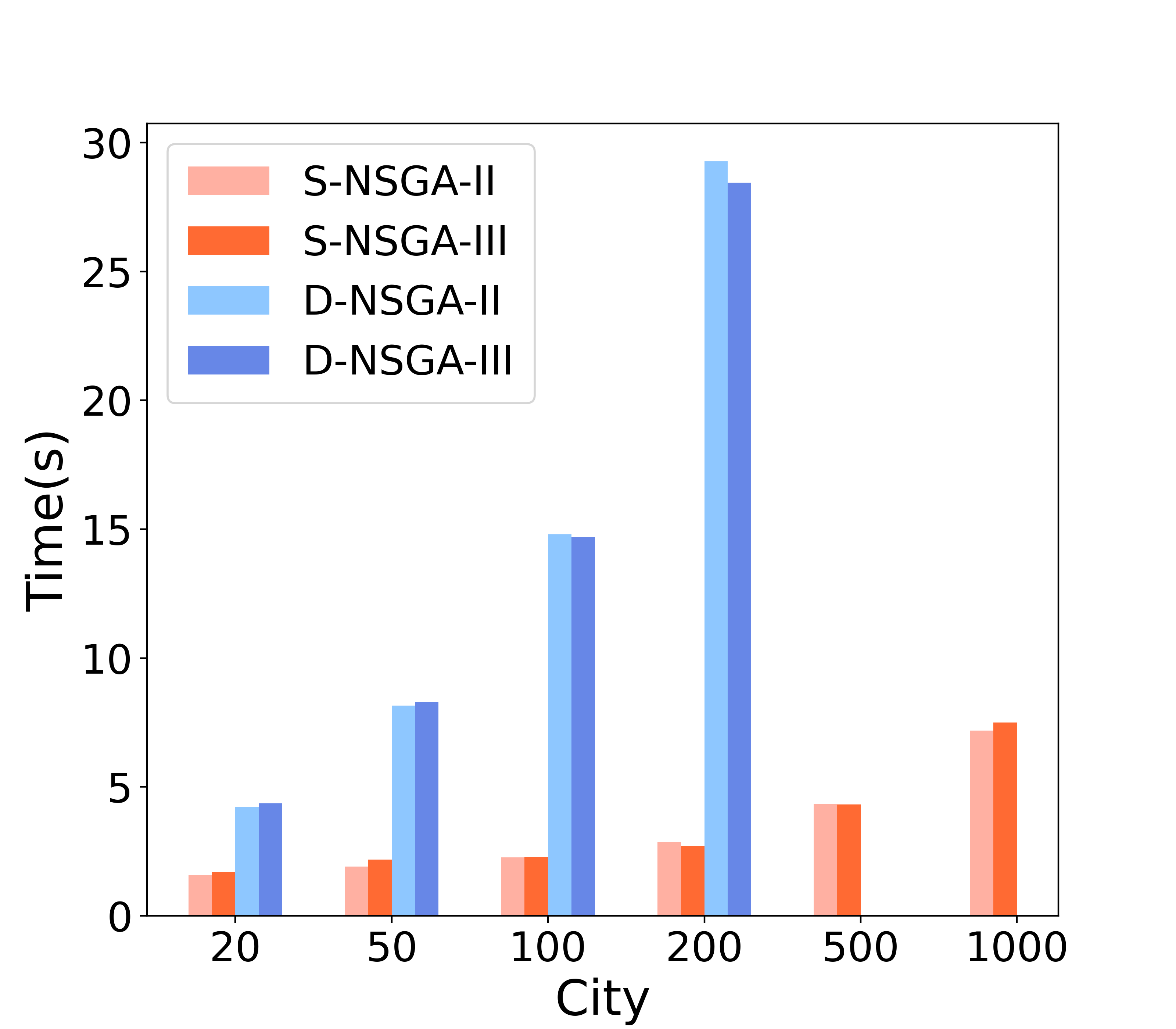}%
}
\hfil
\subfloat[Profits type instances]{\includegraphics[width=1.7in]{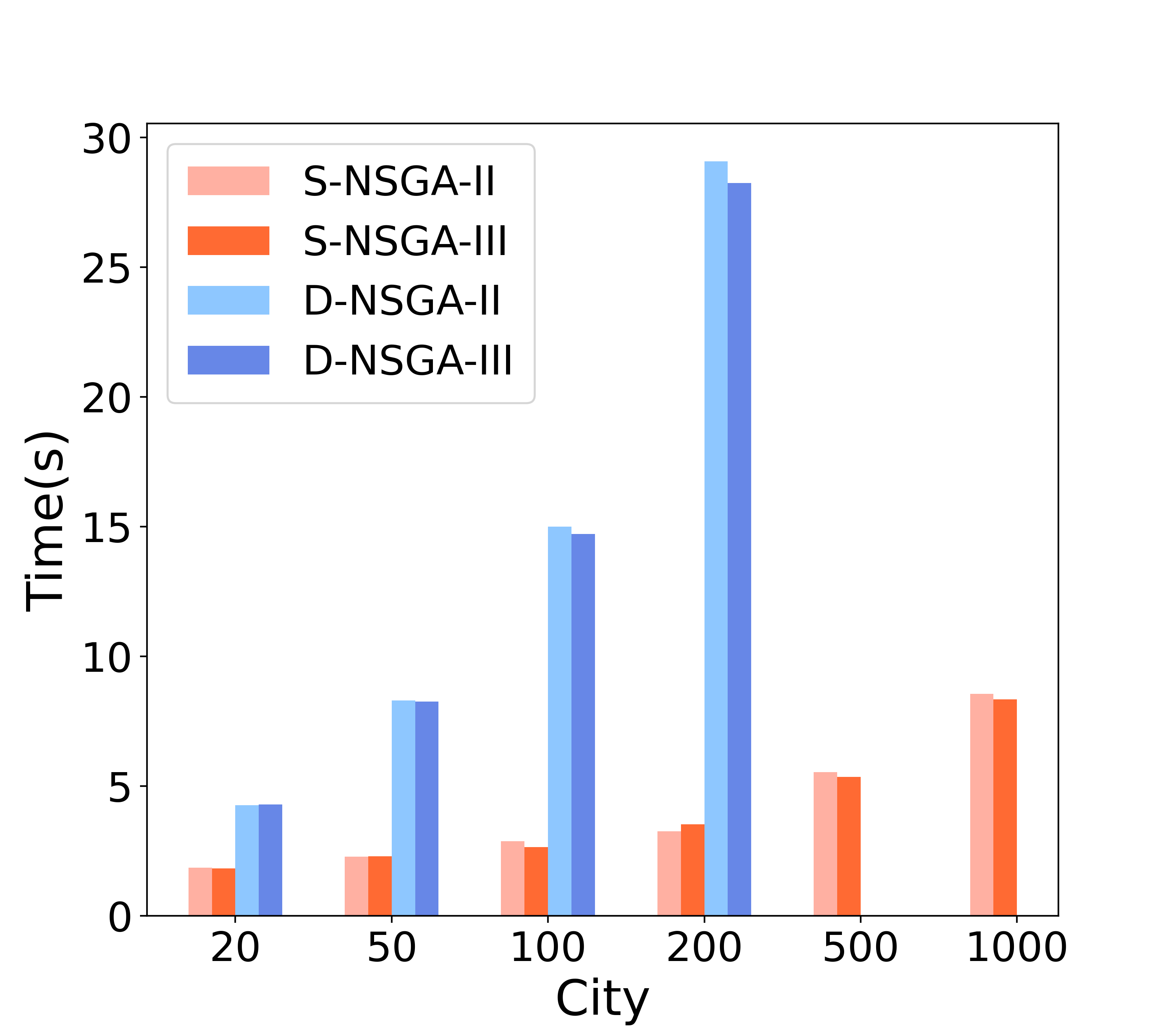}%
}
\caption{Time consumption on mixed and profits type bi-objective OP instances with 20, 50, 100, 200, 500 and 1000 cities. NSGA-II and NSGA-III based on the single-chromosome and double-chromosome coding methods are tested.}
\label{fig:time on coding}
\end{figure}

We can see from these figures that the two coding methods show similar performance on small-scale instances (i.e., instances with 20, 50 and 100 cities). However, as the number of cities increases, the single-chromosome based permutation coding method gradually shows its advantage. In both the mixed type and the profits type instances with 200 cities, algorithms with the single-chromosome permutation coding method outperform those with the double-chromosome coding method. Moreover, since it is difficult for the double-chromosome coding method to guarantee generating feasible solutions during evolution and there is no repair operator being introduced, this coding method loses its efficacy on large-scale instances with 500 and 1000 cities. Different from the double-chromosome coding, solutions generated by the single-chromosome permutation coding are always feasible, which shows more advantages on profits type instances. This is because the single-chromosome permutation coding prefers to visit more cities under the total tour length constraint, resulting in the preference for long tours with large profits as shown in \fref{fig:coding method1}. Due to the higher coding complexity, algorithms with the double-chromosome coding method consume more time in most instances, as shown in \fref{fig:time on coding}. The time consumption gap between these two coding methods becomes larger and larger as the problem size increases. Since better results are consistently obtained from the single-chromosome permutation coding, this coding is used in all algorithms in this study. 

\subsection{Results on Mixed Type Bi-objective OP}
\label{mixed}
\begin{figure}[!ht]
\centering
\subfloat[Compare NSGA-II with our framework on \textbf{Mixed type} instances]{\includegraphics[width=3.3in]{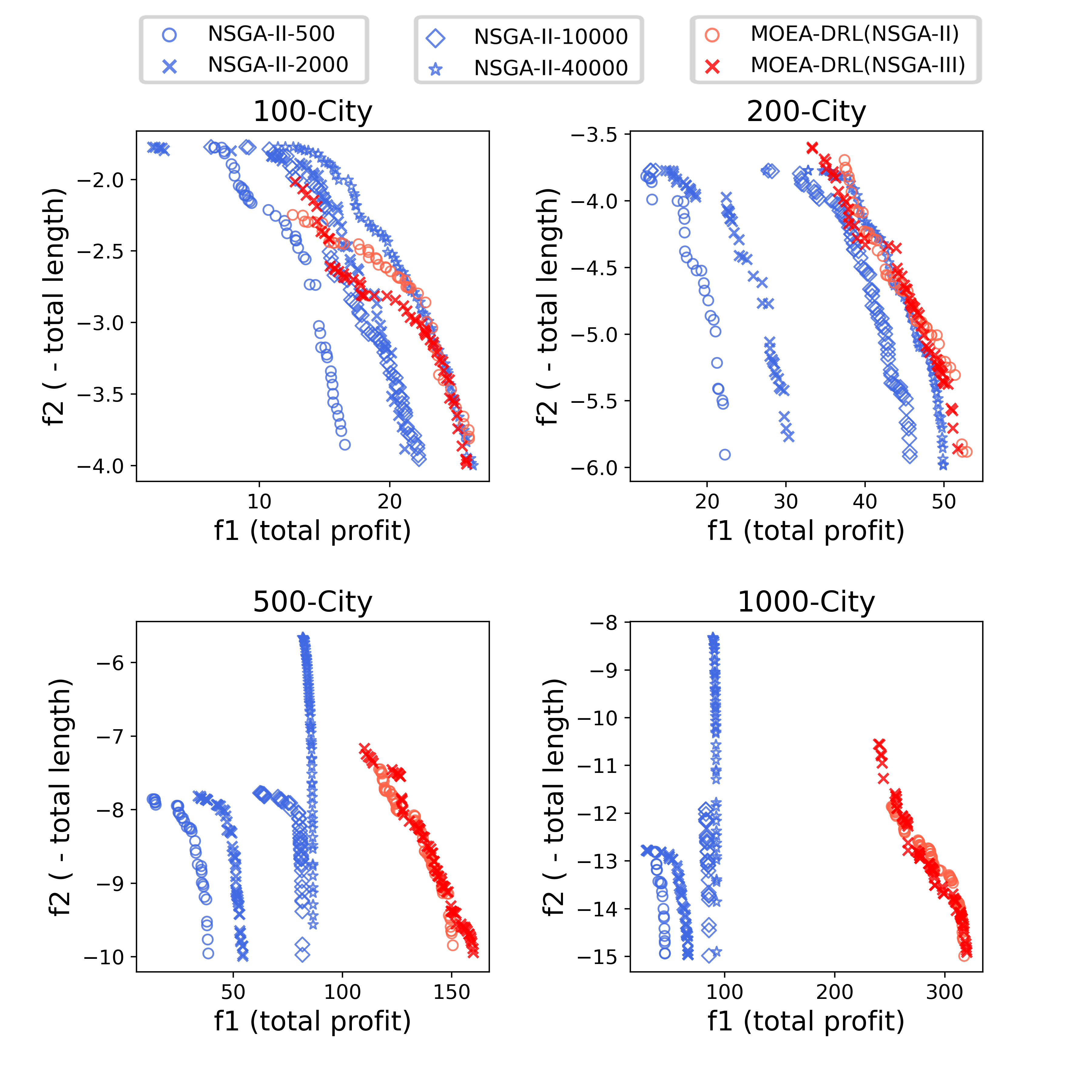}%
}
\hfil
\subfloat[Compare NSGA-III with our framework on \textbf{Mixed type} instances]{\includegraphics[width=3.3in]{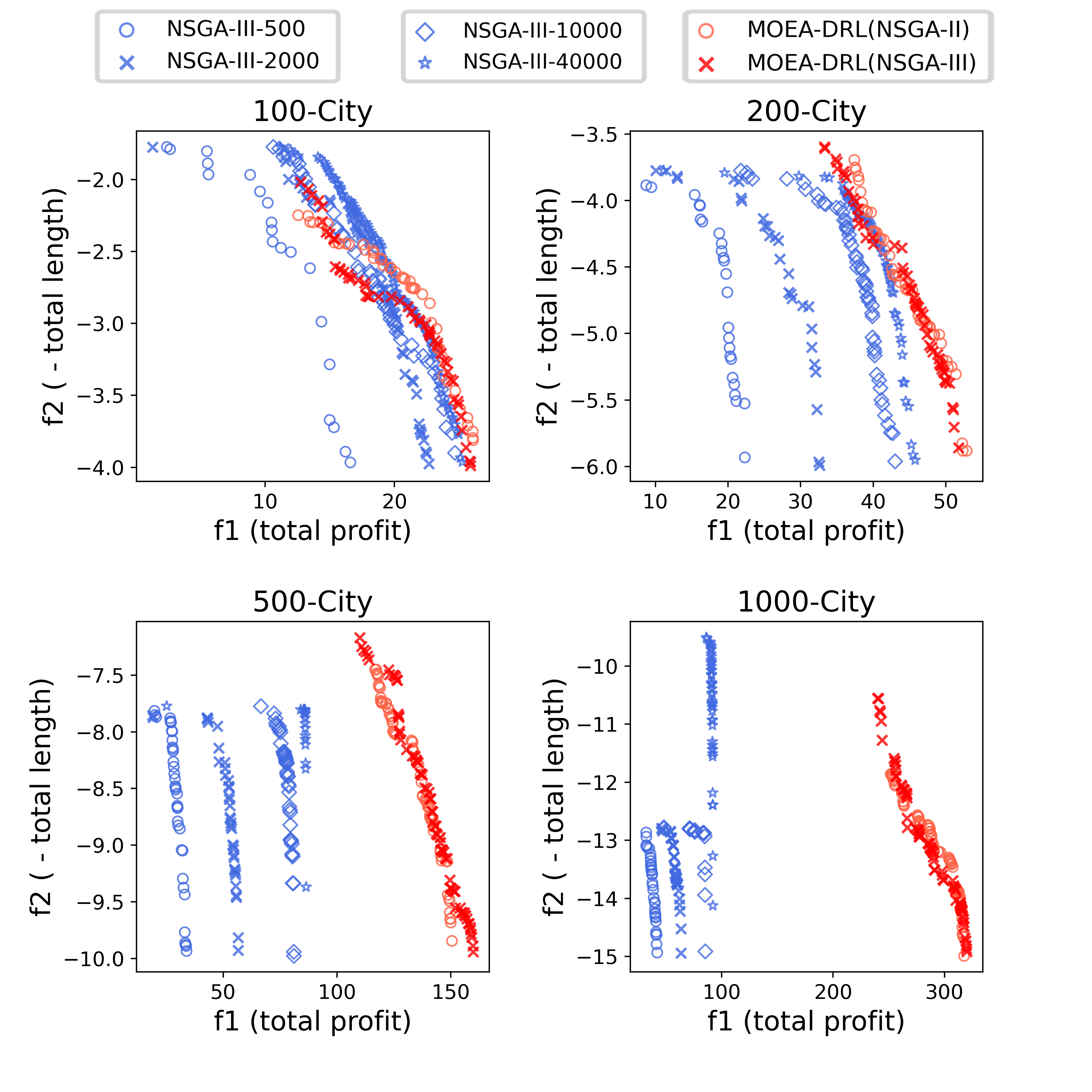}%
}
\caption{Solution sets obtained by MOEA-DRL(NSGA-II), MOEA-DRL(NSGA-III), NSGA-II and NSGA-III. The test instances are randomly generated \textbf{Mixed type} bi-objective OP instances with 100, 200, 500 and 1000 cities.}
\label{fig:mixed type}
\end{figure}

The DRL model trained on 100-city instances is used in the MOEA-DRL framework, which is tested on a single instance for each of 20-, 50-, 100-, 200-, 500- and 1000-city mixed type bi-objective OP. The test instances are generated in the same manner as the setup way in \sref{instance}. Note that MOEA-DRL with NSGA-II and NSGA-III as MOKP solvers are called ``MOEA-DRL(NSGA-II)" and ``MOEA-DRL(NSGA-III)", respectively. NSGA-II and NSGA-III are also directly examined in solving MO-OPs under four different termination conditions, i.e., 500, 2000, 10000 and 40000 generations. 
31 runs for each algorithm on each test instance are conducted. The average HV values and the average running time over these 31 runs are shown in \tref{table:mixed result}. \fref{fig:mixed type} shows the obtained solution set on the instances with 100, 200, 500 and 1000 cities of a single run with the median HV value from those 31 runs. The obtained solution sets for the 20-city and 50-city instances are not presented in \fref{fig:mixed type} since all the competitor algorithms work well on these small-scale instances. 

As shown in \tref{table:mixed result}, all algorithms have similar performance on small-scale instances with 20, 50 and 100 cities. In these instances, NSGA-II and NSGA-III with a large number of generations have slightly higher HV values than MOEA-DRL. However, with the increasing of the problem size, MOEA-DRL gradually shows its advantage over NSGA-II and NSGA-III. Specifically, while NSGA-II and NSGA-III show similar performance to MOEA-DRL(NSGA-II) and MOEA-DRL(NSGA-III) on the 100-city instance, NSGA-II and NSGA-III exhibit an obviously inferior performance compared with our framework on the 500- and 1000-city instances. It is not likely that NSGA-II and NSGA-III can achieve comparable performance to MOEA-DRL by increasing the number of generations, see \fref{fig:mixed type}, since the progress from 10,000 generations to 40,000 generations is minor.

The total number of examined solutions in the MOEA-DRL framework is much smaller than that of MOEAs. While the population size in all the competitor algorithms is 100, the maximum number of generations in our framework is only 20 (and up to 40,000 in NSGA-II and NSGA-III). When these algorithms are compared under the same running time (excluding the training time for the DYPN model), it is clear that the MOEA-DRL framework outperforms MOEAs. There is little difference between NSGA-II and NSGA-III on both evaluation quality and running time in \tref{table:mixed result}. MOEA-DRL(NSGA-III) shows slightly better performance than MOEA-DRL(NSGA-II). It is worth noting that in MOEA-DRL the DYPN model has to be pre-trained. However, this is acceptable since in most scenarios model training can be offline or conducted in advance. Once DYPN is trained, the MOEA-DRL can be much more effective than MOEAs on large-scale instances as shown in \fref{fig:mixed type}.

\begin{table*}[htbp]
\setlength\tabcolsep{4pt}
\centering
\caption{Average HV values and running time obtained by NSGA-II, NSGA-III and MOEA-DRL. The test instances are \textbf{Mixed type} bi-objective OP instances. The best HV value is marked in gray background.}
\begin{tabular}{cccccccccccccc}
\hline
                           &                                      & \multicolumn{2}{c}{\textbf{20-city}}          & \multicolumn{2}{c}{\textbf{50-city}}           & \multicolumn{2}{c}{\textbf{100-city}}          & \multicolumn{2}{c}{\textbf{200-city}}           & \multicolumn{2}{c}{\textbf{500-city}}           & \multicolumn{2}{c}{\textbf{1000-city}}           \\ \cline{3-14} 
                           & \multirow{-2}{*}{\textbf{Examined Solutions}} & \textbf{HV}                 & \textbf{Time/s} & \textbf{HV}                  & \textbf{Time/s} & \textbf{HV}                  & \textbf{Time/s} & \textbf{HV}                   & \textbf{Time/s} & \textbf{HV}                   & \textbf{Time/s} & \textbf{HV}                    & \textbf{Time/s} \\ \hline
\textbf{NSGA-II-500}        & 5.0E+04                              & \cellcolor[HTML]{C0C0C0}4.3 & 1.6             & 19.0                         & 1.9             & 29.6                         & 2.3             & 42.6                          & 2.9             & 71.6                          & 4.3             & 93.4                           & 7.2             \\
\textbf{NSGA-II-2000}       & 2.0E+05                              & \cellcolor[HTML]{C0C0C0}4.3 & 6.3             & 22.9                         & 8.0             & 39.6                         & 9.4             & 58.5                          & 12.7            & 109.2                         & 22.5            & 133.8                          & 33.5            \\
\textbf{NSGA-II-10000}      & 1.0E+06                              & \cellcolor[HTML]{C0C0C0}4.3 & 34.4           & 23.0                           & 43.6           & 39.0                           & 51.7           & 91.4                         & 81.3           & 178.4                         & 145.4          & 261.3                          & 217.8          \\
\textbf{NSGA-II-40000}      & 4.0E+06                              & \cellcolor[HTML]{C0C0C0}4.3 & 134.3           & \cellcolor[HTML]{C0C0C0}23.6 & 173.6           & \cellcolor[HTML]{C0C0C0}47.9 & 224.2           & 100.5                         & 379.2           & 371.7                         & 670.5           & 615.0                          & 912.4           \\
\textbf{NSGA-III-500}       & 5.0E+04                              & \cellcolor[HTML]{C0C0C0}4.3 & 1.7             & 17.7                         & 2.2             & 27.7                         & 2.3             & 41.4                          & 2.7             & 66.5                          & 4.3             & 81.8                           & 7.5             \\
\textbf{NSGA-III-2000}      & 2.0E+05                              & \cellcolor[HTML]{C0C0C0}4.3 & 6.9             & 21.6                         & 8.4             & 41.4                         & 10.8            & 63.6                          & 12.9            & 112.9                         & 21.7            & 132.0                          & 34.7            \\
\textbf{NSGA-II-10000}      & 1.0E+06                              & \cellcolor[HTML]{C0C0C0}4.3 & 34.6            & 22.8                         & 44.4            & 42.4                         & 54.4            & 85.5                         & 86.0           & 175.1                         & 147.8          & 187.6                          & 212.3          \\
\textbf{NSGA-II-40000}      & 4.0E+06                              & \cellcolor[HTML]{C0C0C0}4.3 & 137.9           & 23.3                         & 179.2           & 45.7                         & 234.6           & 93.0                          & 362.2           & 190.3                         & 673.7           & 503.5                          & 944.5           \\
\textbf{MOEA-DRL(NSGA-II)}  & 2.0E+03                              & 3.7                         & 31.9            & 19.3                         & 50.6            & 39.0                         & 80.6            & 105.6                         & 148.9           & 352.5                         & 461.1           & 924.9                          & 930.3           \\
\textbf{MOEA-DRL(NSGA-III)} & 2.0E+03                              & 4.0                         & 32.5            & 19.9                         & 53.1            & 40.4                         & 76.0            & \cellcolor[HTML]{C0C0C0}107.4 & 151.2           & \cellcolor[HTML]{C0C0C0}391.8 & 462.3           & \cellcolor[HTML]{C0C0C0}1236.9 & 896.8           \\ \hline
\end{tabular}
\label{table:mixed result}%
\end{table*}

\subsection{Results on Profits Type Bi-objective OP}
\label{profits}

\begin{table*}[!ht]
	\setlength\tabcolsep{3.5pt}
	\centering
	\caption{Average HV values and running time obtained by NSGA-II, NSGA-III and MOEA-DRL. The test instances are \textbf{Profits type} bi-objective OP instances. The best HV value is marked in gray background.}
	\begin{tabular}{cccccccccccccc}
		\hline
		\multicolumn{1}{l}{}       &                                      & \multicolumn{2}{c}{\textbf{20-city}}           & \multicolumn{2}{c}{\textbf{50-city}}            & \multicolumn{2}{c}{\textbf{100-city}}           & \multicolumn{2}{c}{\textbf{200-city}}            & \multicolumn{2}{c}{\textbf{500-city}}             & \multicolumn{2}{c}{\textbf{1000-city}}            \\ \cline{3-14} 
		\multicolumn{1}{l}{}       & \multirow{-2}{*}{\textbf{Examined Solutions}} & \textbf{HV}                  & \textbf{Time/s} & \textbf{HV}                   & \textbf{Time/s} & \textbf{HV}                   & \textbf{Time/s} & \textbf{HV}                    & \textbf{Time/s} & \textbf{HV}                     & \textbf{Time/s} & \textbf{HV}                     & \textbf{Time/s} \\ \hline
		\textbf{NSGA-II-500}        & 5.0E+04                              & 30.1                         & 1.9             & 130.8                         & 2.3             & 248.6                         & 2.9             & 583.7                          & 3.3             & 1317.9                          & 5.5             & 2425.3                          & 8.6             \\
		\textbf{NSGA-II-2000}       & 2.0E+05                              & 35.1                         & 7.4             & 186.5                         & 9.6             & 392.0                         & 12.0            & 1129.5                         & 17.6            & 2163.5                          & 25.5            & 4130.0                          & 42.3            \\
		\textbf{NSGA-II-10000}      & 1.0E+06                              & 35.1                         & 38.5            & 234.4                         & 56.5            & 442.7                         & 70.0            & 1185.6                         & 101.5           & 4198.8                          & 176.8           & 4282.0                          & 230.5           \\
		\textbf{NSGA-II-40000}      & 4.0E+06                              & 36.1                         & 154.2           & 245.2                         & 203.3           & 601.9                         & 286.8           & 1811.5                         & 395.8           & 5460.2                          & 675.7           & 6705.9                          & 921.1           \\
		\textbf{NSGA-III-500}       & 5.0E+04                              & 33.5                         & 1.8             & 152.1                         & 2.3             & 215.6                         & 2.7             & 609.5                          & 3.5             & 1427.5                          & 5.4             & 1961.5                          & 8.4             \\
		\textbf{NSGA-III-2000}      & 2.0E+05                              & 29.5                         & 7.8             & 183.6                         & 10.2            & 421.8                         & 13.0            & 926.3                          & 16.7            & 2800.1                          & 28.9            & 3342.0                          & 40.2            \\
		\textbf{NSGA-II-10000}      & 1.0E+06                              & 33.8                         & 39.2            & 203.2                         & 50.8            & 377.6                         & 68.4            & 1662.3                         & 102.3           & 3783.1                          & 168.9           & 4191.5                          & 228.0           \\
		\textbf{NSGA-II-40000}      & 4.0E+06                              & 33.8                         & 155.3           & 233.7                         & 204.4           & 421.7                         & 299.6           & 2037.2                         & 422.5           & 4203.4                          & 745.5           & 6209.0                          & 931.6           \\
		\textbf{MOEA-DRL(NSGA-II)}  & 2.0E+03                              & \cellcolor[HTML]{C0C0C0}36.3 & 42.5            & \cellcolor[HTML]{C0C0C0}253.9 & 60.6            & 576.1                         & 84.8            & \cellcolor[HTML]{C0C0C0}2341.1 & 159.1           & \cellcolor[HTML]{C0C0C0}20071.6 & 499.7           & \cellcolor[HTML]{C0C0C0}78677.8 & 965.0           \\
		\textbf{MOEA-DRL(NSGA-III)} & 2.0E+03                              & 34.8                         & 45.3            & 253.0                         & 58.2            & \cellcolor[HTML]{C0C0C0}604.2 & 85.6            & 2188.0                         & 158.3           & 17197.6                         & 487.8           & 74425.4                         & 1034.6          \\ \hline
	\end{tabular}
	\label{table:profits result}%
\end{table*}

\begin{figure}[htbp]
\centering
\subfloat[Compare NSGA-II with our framework on \textbf{Profits type} instances]{\includegraphics[width=3.3in]{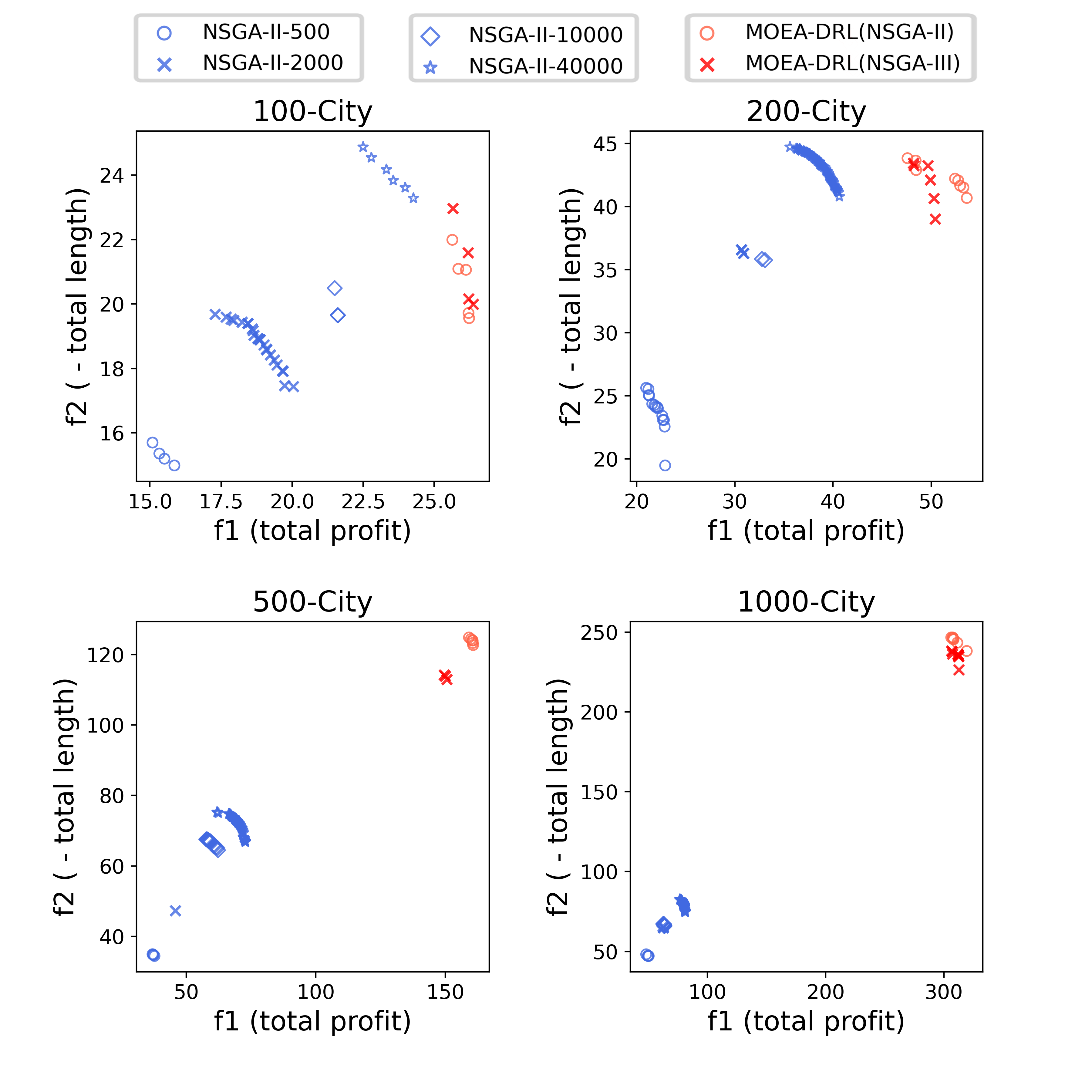}%
}
\hfil
\subfloat[Compare NSGA-III with our framework on \textbf{Profits type} instances]{\includegraphics[width=3.3in]{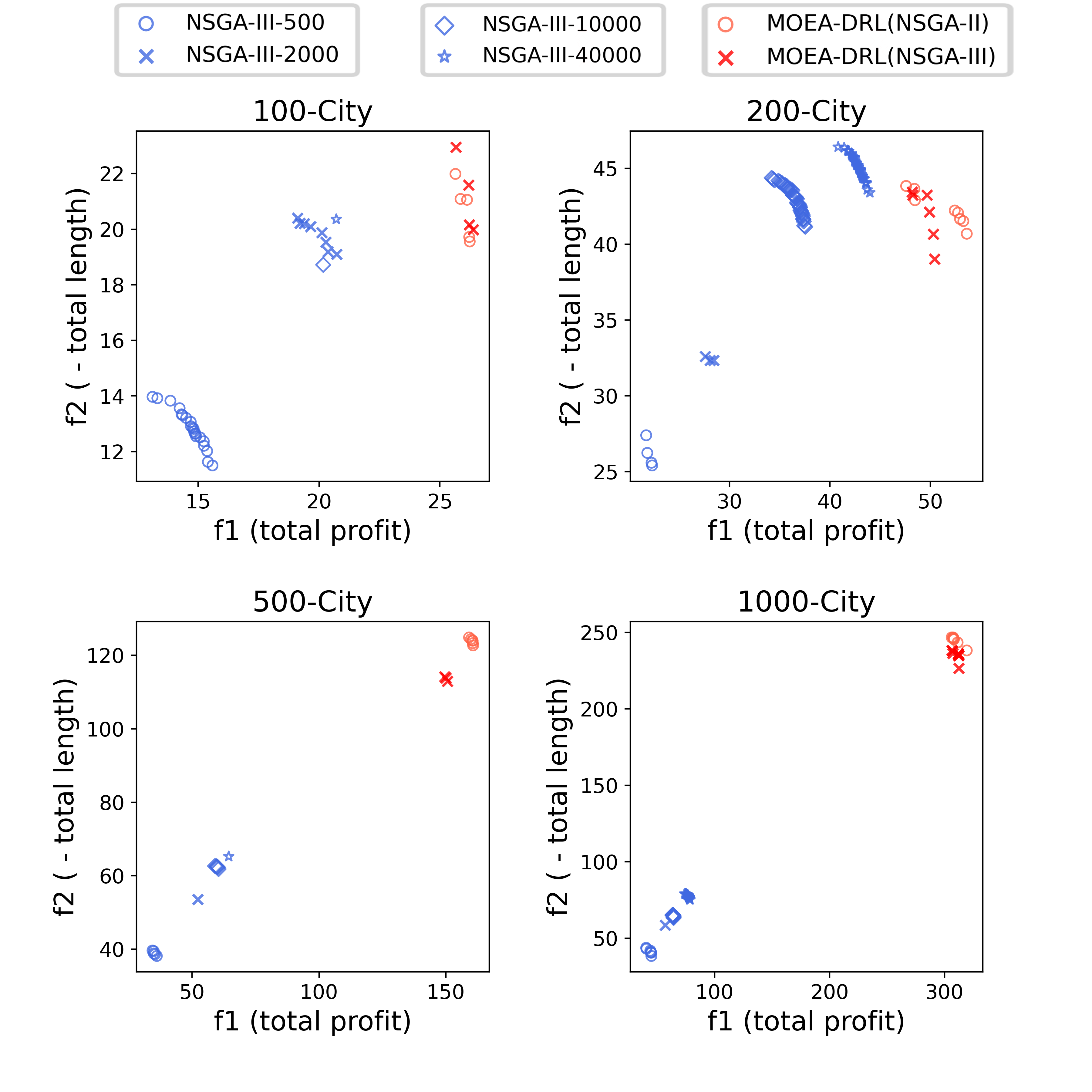}%
}
\caption{Solution sets obtained by MOEA-DRL(NSGA-II), MOEA-DRL(NSGA-III), NSGA-II and NSGA-III. The test instances are randomly generated \textbf{Profits type} bi-objective OP instances with 100, 200, 500 and 1000 cities.}
\label{fig:profits type}
\end{figure}

In this subsection, we test our framework on a single instance for each of 20-, 50-, 100-, 200-, 500- and 1000-city profits type bi-objective OP. The test instances are generated in the same manner as in \sref{instance}. 31 runs for each algorithm on each test instance are conducted. The average HV values and the average running time over these 31 runs are shown in \tref{table:profits result}. \fref{fig:profits type} shows the obtained solution set on the instances with 100, 200, 500 and 1000 cities of a single run with the median HV value from those 31 runs. The solution sets on 20- and 50-city instances are not shown since all algorithms perform comparably.

It is clear that the number of the obtained solutions in \fref{fig:profits type} on the profits type instances is much smaller than that on the mixed type instances in \fref{fig:mixed type}. This is explained as follows. In the mixed type instances, the total profit and the total tour length are clearly conflicting. As a result, a large number of widely-distributed no-dominated solutions are obtained in \fref{fig:mixed type}. In profits type instances, two profits at each city are randomly generated. Thus, the two profit objectives are not necessarily conflicting. Moreover, under the total tour length constraint, each algorithm tries to visit cities as many as possible. For these reasons, only a few non-dominated solutions are obtained in \fref{fig:profits type} for profits type instances, and their distribution is not diversified. However, profits type bi-objective OPs are still practically important \cite{schilde2009metaheuristics}.

As shown in \fref{fig:profits type}, although the increase in the number of generations gradually improves the convergence of solutions in NSGA-II and NSGA-III, the performance of NSGA-II and NSGA-III is far weaker than that of our framework, especially in large-scale instances. HV-based comparison results in \tref{table:profits result} also confirm this performance difference. That is, in \tref{table:profits result}, MOEA-DRL(NSGA-II) and MOEA-DRL(NSGA-III) significantly outperform NSGA-II and NSGA-III. The performance gap becomes larger and larger with the increase of the problem size. Under the same computation time, our framework outperforms NSGA-II and NSGA-III on all problem instances. 

\subsection{Results on Three-objective OP}
\label{3D}
\begin{figure}
	\centering
	\includegraphics[width=1\linewidth]{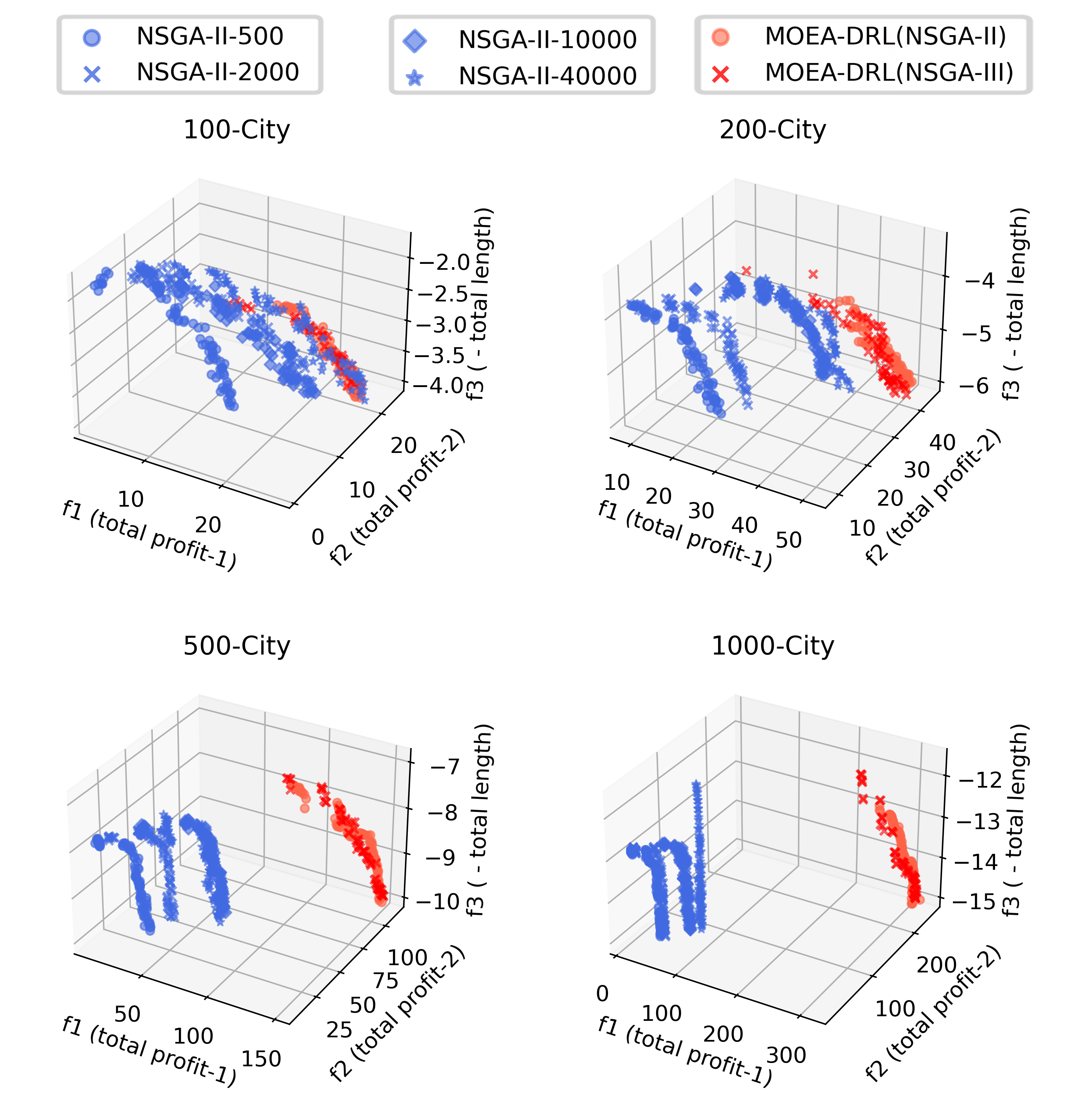}
	\caption{Solution sets obtained by MOEA-DRL(NSGA-II), MOEA-DRL(NSGA-III) and NSGA-II. The test instances are randomly generated \textbf{three-objective} OP instances with 100, 200, 500 and 1000 cities.}
\label{fig:3objs}
\end{figure}

In this subsection, we test our framework on three-objective OP instances. The three objectives are to maximize the two types of total profits and to minimize the tour length. Our framework is tested on a single three-objective OP instance for each of 50-, 100-, 200- and 500-city problems. All these instances are generated in the same manner as the setup way in \sref{instance}. We examined both NSGA-II and NSGA-III under four termination conditions--- 500, 2000, 10000 and 40000 generations (as competitor algorithms), and obtained similar results. Thus, this subsection only presents results obtained by NSGA-II under 500, 2000, 10000 and 40000 generations. The average HV values over 31 runs are shown in \tref{table:3objs result} for each algorithm on each problem instance. \fref{fig:3objs} shows the obtained solution set of a single run with the median HV value from those 31 runs.

While NSGA-II with 40,000 generations shows slight superior performance on the 100-city test instance, MOEA-DRL significantly outperforms NSGA-II on the 200- and 500-city instances. Moreover, the gap of the average HV values between MOEA-DRL and NSGA-II becomes lager and larger when the problem size increases.

\begin{table}[htbp]
\setlength\tabcolsep{2pt}
\centering
\caption{Average HV values obtained by NSGA-II and MOEA-DRL. The test instances are \textbf{three-objective} OP instances. The best HV value is marked in gray background.}
\begin{tabular}{cccccc}
\hline
\textbf{HV}                & \textbf{Solutions} & \textbf{100-city}             & \textbf{200-city}              & \textbf{500-city}               & \textbf{1000-city}               \\ \hline
\textbf{NSGA-II-500}        & 5.0E+04                   & 265.3                         & 676.2                          & 1613.5                          & 3138.1                           \\
\textbf{NSGA-II-2000}       & 2.0E+05                   & 512.8                         & 1099.6                         & 3250.8                          & 4401.9                           \\
\textbf{NSGA-II-10000}      & 1.0E+06                   & 508.2                         & 2392.4                         & 7759.5                          & 10338.1                          \\
\textbf{NSGA-II-40000}      & 4.0E+06                   & \cellcolor[HTML]{C0C0C0}788.2 & 2803.3                         & 7909.6                          & 21768.4                          \\
\textbf{MOEA-DRL(NSGA-II)}  & 2.0E+03                   & 685.0                         & 3403.8                         & 39745.8                         & 173910.2                         \\
\textbf{MOEA-DRL(NSGA-III)} & 2.0E+03                   & 673.9                         & \cellcolor[HTML]{C0C0C0}4051.8 & \cellcolor[HTML]{C0C0C0}40811.4 & \cellcolor[HTML]{C0C0C0}206401.7 \\ \hline
\end{tabular}
\label{table:3objs result}%
\end{table}

\subsection{Training on Different Number of Cities}
\label{different training}

\begin{figure}
	\centering
	\includegraphics[width=1\linewidth]{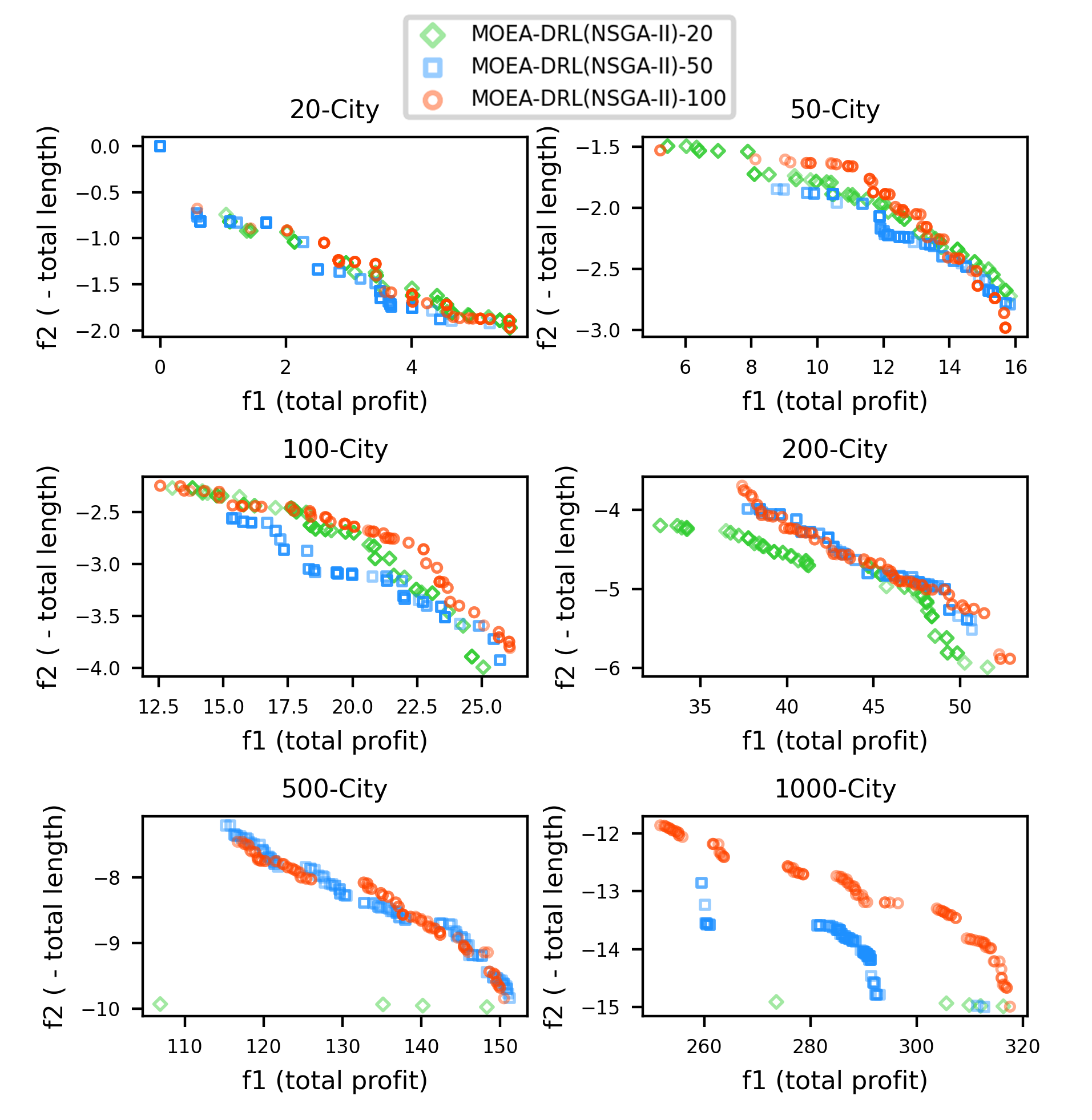}
	\caption{Performance comparison of MOEA-DRL(NSGA-II) trained on 20-, 50- and 100-city instances. \textbf{Mixed type} bi-objective OP instances with 20, 50, 100, 200, 500 and 1000 cities are tested.}
	\label{fig:different training}
\end{figure}

In the previous subsections, the DRL model trained by 100-city TSP instances is always used in the MOEA-DRL framework on test instances with 20-1000 cities. To further examine the generalization ability of MOEA-DRL, we examine the use of DRL models trained on TSP instances with 20, 50 and 100 cities, respectively. Test instances in this subsection are 20-, 50-, 100-, 200-, 500- and 1000-city mixed type bi-objective OP instances, which are generated in the same manner as in \sref{instance}. The population size is set as 100 and the termination condition is 20 generations in all the three implementations of MOEA-DRL (i.e., with three DRL models trained on TSP instances of a different size--- 20-, 50-, and 100-city instances). 31 runs of each implementation are conducted for each test instance. \fref{fig:different training} shows the obtained solution set of a single run with the median performance from those 31 runs. 

Based on \fref{fig:different training}, when the test instance size increases to 200, the performance of the model trained on 20-city instances starts to deteriorate. It finds only a small number of non-dominated solutions with low quality on 500- and 1000-city instances. Similarly, the model trained on 50-city instances shows weakness when the instance size increases to 1000. Whereas the MOEA-DRL framework shows strong generalization ability to unseen instances, a large gap in the problem size between training instances and testing instances leads to clear performance degradation.

\subsection{Summary of the Results}
\label{summary}

The experimental results can be summarized as follows:
\begin{itemize}
\item The introduction of dynamic information to the DRL model DYPN improves both of the convergence speed and solution quality. 
\item The single-chromosome based permutation coding method is more effective (i.e., faster evolution speed and higher solution quality) for MO-OPs compared to the double-chromosome coding method.
\item For all the mixed type bi-objective instances, the profits type bi-objective instances and the three-objective OP instances, the MOEA-DRL framework shows clear advantages on solution quality, especially for large-scale instances.
\item The MOEA-DRL framework shows strong generalization ability, which performs well on 200-, 500- and 1000-city instances even when the DRL model is trained on 100-city instances. 
\end{itemize}

\section{Conclusion}
\label{conclusion}
In this study, we proposed a new idea of solving MO-OPs by problem decomposition. More specifically, we decomposed a MO-OP into a MOKP and a TSP, and proposed a hybrid optimization framework MOEA-DRL which hybridizes a MOEA and a DRL to solve MOKP and TSP, respectively. The MOEA-DRL framework was evaluated through computational experiments on randomly generated mixed type bi-objective instances, profits type bi-objective instances and three-objective instances. Experimental results showed that the MOEA-DRL framework greatly outperforms NSGA-II and NSGA-III, especially, for large-scale problem instances. It was also shown that the MOEA-DRL framework has high generalization ability. For example, it worked well on 1000-city instances even when the DRL model was trained on 100-city instances. Whereas the training of a DRL model takes hours, the trained model can be used for all test instances with different objectives and different number of cities.

One future research direction is to improve the performance of the MOEA-DRL framework by using more advanced MOEA and DRL. Especially, its efficiency improvement is important with respect to both the DRL training and the MOEA search. It is also a promising research direction to implement multi-objective algorithms based on the proposed MOEA-DRL framework for various variants of MO-OPs in the future.


%



\ifCLASSOPTIONcaptionsoff
  \newpage
\fi



\bibliographystyle{IEEEtran}
\bibliography{IEEEfull}




\end{document}